\documentclass{article}
\usepackage{jfrExamplee}
\usepackage{graphicx}
\usepackage{apalike}
\usepackage{setspace}

\usepackage{todonotes}

\usepackage{siunitx} 
\usepackage{tensor} 
\usepackage{amsmath}

\usepackage{subfig}

\newcommand{\abs}[1]{\left\lvert#1\right\rvert} 

\DeclareMathOperator{\mean}{mean}
\DeclareMathOperator{\std}{std}

\title{Tethered Aerial Visual Assistance}

\author{
Xuesu Xiao\thanks{Xuesu Xiao was a PhD student at Texas A\&M University, where the majority of this work was conducted.} \\
Department of Computer Science\\
University of Texas at Austin\\
2317 Speedway, Austin, TX 78712 \\
\texttt{xiao@cs.utexas.edu} \\
\And
Jan Dufek \\
Department of Computer Science and Engineering \\
Texas A\&M University \\
3112 TAMU, College Station, TX 77843 \\
\texttt{dufek@tamu.edu} \\
\And
Robin R. Murphy \\
Department of Computer Science and Engineering \\
Texas A\&M University \\
3112 TAMU, College Station, TX 77843 \\
\texttt{robin.r.murphy@tamu.edu} \\
}

\begin{document}

\maketitle

\begin{abstract}
In this paper, an autonomous tethered Unmanned Aerial Vehicle (UAV) is developed into a visual assistant in a marsupial co-robots team, collaborating with a tele-operated Unmanned Ground Vehicle (UGV) for robot operations in unstructured or confined environments. 
These environments pose extreme challenges to the remote tele-operator due to the lack of sufficient situational awareness, mostly caused by the unstructuredness and confinement, stationary and limited field-of-view and lack of depth perception from the robot's onboard cameras.
To overcome these problems, a secondary tele-operated robot is used in current practices, who acts as a visual assistant and provides external viewpoints to overcome the perceptual limitations of the primary robot's onboard sensors. However, a second tele-operated robot requires extra manpower and teamwork demand between primary and secondary operators. The manually chosen viewpoints tend to be subjective and sub-optimal. 
Considering these intricacies, we develop an autonomous tethered aerial visual assistant in place of the secondary tele-operated robot and operator, to reduce human robot ratio from 2:2 to 1:2. Using a fundamental viewpoint quality theory, a formal risk reasoning framework, and a newly developed tethered motion suite, our visual assistant is able to autonomously navigate to good-quality viewpoints in a risk-aware manner through unstructured or confined spaces with a tether. 
The developed marsupial co-robots team could improve tele-operation efficiency in nuclear operations, bomb squad, disaster robots, and other domains with novel tasks or highly occluded environments, by reducing manpower and teamwork demand, and achieving better visual assistance quality with trustworthy risk-aware motion. 
\end{abstract}

\section{Introduction}
Despite the progressive advances in robot autonomy, tele-operated robots are still preferred in terms of their predictability, reliability, robustness, flexibility, and versatility. Therefore they are still being widely used in many DDD (Dangerous, Dirty, and Dull) scenarios, where human presence is extremely difficult or impossible. This is especially the case during actual field deployment in unstructured or confined environments, due to those environments' mission-critical task execution and current technological limitations. 

\begin{figure}[h]
\centering
	\includegraphics[scale=0.4]{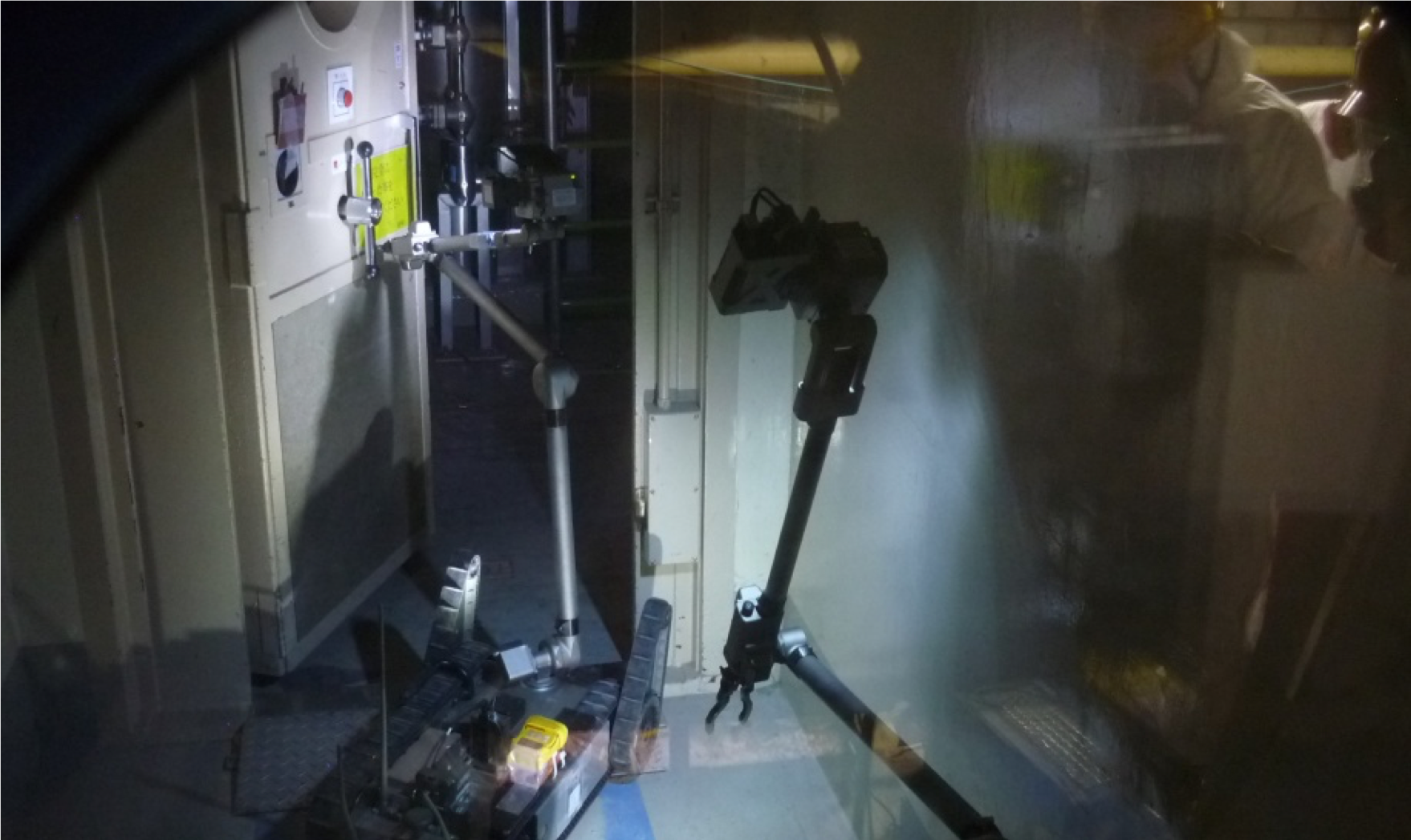}
	\caption{Two iRobot PackBots Working Together to Open a Door during the Fukushima Daiichi Nuclear Accident (Image Courtesy: JAEA)}
	\label{fig::2packbots}
\end{figure}

Still being widely used as an effective approach to leverage current technologies and actual field demand, tele-operation, as a means of projecting human presence, also requires instant and comprehensive feedback and understanding of the remote environment for the tele-operator. Such situational awareness is always limited by only using onboard sensors, such as relatively stationary and limited field of view and lack of depth perception from the tele-operated robot's onboard cameras. To overcome this limitation, robotic deployment for nuclear operations, bomb squad, disaster robots, and other domains with novel tasks or highly occluded environments uses two robots, a primary and a secondary that acts as a visual assistant to overcome the perceptual limitations of the onboard sensors by providing an external viewpoint. In Fukushima Daiichi nuclear power plant accident in 2011, for example, tele-operated robots were used in pairs from the beginning of the response to reduce the time it took to accomplish a task \cite{murphy2014disaster}. iRobot PackBots were used to conduct radiation surveys and read dials inside the plant facility, where the second PackBot provided camera views of the first robot in order to manipulate door handles, valves, and sensors faster (Fig. \ref{fig::2packbots}). QinetiQ Talon UGVs let operators see if their tele-operated Bobcat end loader bucket had scraped up a full load of dirt to deposit over radioactive materials. Since then, the use of two robots to perform a single task has been formally acknowledged as a best practice for decommissioning tasks, e.g., cutting and removing a section of irradiated pipe. At the Deepwater Horizon spill, as example in another regime, a second Underwater Remotely Operated Vehicle (ROV) was used to help position the primary ROV so that it could insert the cap to stop the leaking oil \cite{murphy2014disaster}. 

Tele-operating a second robot as visual assistant, however, requires extra human operators, who may also choose sub-optimal viewpoints based only on personal experience. On top of the task and perceptual demands of tele-operation, extra teamwork demand \cite{wickens2008multiple} is required for the communication between the primary and secondary operators. The Japanese Atomic Energy Agency (JAEA) has reported to us through our memorandum of understanding for cooperative research on disaster robotics that operators constantly try to avoid using a second robot. The two sets of robot operators find it difficult to coordinate with the other robot in order to get and maintain the desired view but a single operator becomes frustrated trying to operate both robots. However, two robots are better than one. In 2014, an iRobot Warrior costing over \$500K was damaged due to inability to see that it was about to perform an action it could not successfully complete. The experienced operator had declined to use a second robot. Not only was this a direct economic loss, the 150kg robot was too heavy to be removed without being dismantled and thus cost other robots time and increased their risk as they have to navigate around the carcass until another robot could be modified to dismantle it. Incidents happened even when both robots were used: during Deepwater Horizon spill response, after the insertion of the cap, the primary ROV bumped into the observing ROV and careened into the cap \cite{murphy2014disaster}. The cap had to be removed, repaired, and reinstalled, adding several days to the overall mitigation.

To address these issues, in this research, a marsupial co-robots team is developed, with one tele-operated ground robot and an autonomous tethered aerial visual assistant to replace the secondary robot and its operators, reducing the human robot ratio from 2:2 to 1:2. The primary ground robot tele-operator's situational awareness is maintained by the autonomous assistant's visual feedback streamed from the optimal viewpoint for the particular tele-operation task. The autonomous visual assistance is realized through a fundamental viewpoint quality theory, a formal risk reasoning framework and risk-aware planner, and a tethered aerial motion suite, including tether-based localization, motion primitives, and contacts planning. We also demonstrate the marsupial co-robots team through actual field deployment in multiple physical environments. 

The remainder of this article is organized as follows: Sec. \ref{sec::related_work} provides related work regarding UGV-UAV team, visual assistance for tele-operation, robot motion risk, and tethered aerial locomotion. Sec. \ref{sec::team} presents the heterogeneous co-robots team. Sec. \ref{sec::viewpoint_quality} presents a fundamental viewpoint quality theory based on the cognitive science concept of Gibsonian affordances. Sec. \ref{sec::risk-awareness} introduces a formal risk reasoning framework and a risk-aware planner to navigate robots in unstructured or confined environments with maximum safety. Sec. \ref{sec::tethererd_motion} provides a new tethered motion suite for tethered aerial vehicles to negotiate with complex workspaces. Sec. \ref{sec::system_demonstration} presents demonstration of the developed system in different physical environments. Sec. \ref{sec::conclusion} concludes the paper and discusses future work.

\section{Related Work}
\label{sec::related_work}
This section reviews related work in terms of UGV-UAV collaboration, visual assistance for tele-operation, robot motion risk, and aerial robots locomoting with tether. 

\subsection{UGV-UAV Team}
The stable chassis of UGVs allows higher, more reliable and durable payload capacity and can thus represent humans to actuate upon the real world, while UAVs' superior mobility and workspace coverage make them capable of providing enhanced situational awareness \cite{murphy2016two}. Teaming of the two types of platforms can overcome one's weaknesses by acquiring other's strength. For example, UAV was used to cover a wide area, geolocate a dynamic target and share with UGV to pursue \cite{cheung2008uav}, or stabilizing UGVs into a guarding formation, and using UAVs to scan the enclosed regions to detect target \cite{tanner2007switched}. Air-ground multi-robot teams were deployed in order to increase situational awareness, achieve cooperative sensing, and construct radio maps to keep team connectivity \cite{chaimowicz2005deploying}, or to detect and fight wild fire \cite{phan2008cooperative}. In this category, the UAV flied in outdoor open spaces without the existence of any obstacles. 

A more relevant category was to assist UGV's task execution by augmenting UGV's perception from UAV's onboard camera , such as ``an eye in the sky'' for UGV localization \cite{chaimowicz2004experiments}, providing stationary third person view for construction machine \cite{kiribayashi2018design}, improving navigation in case of GPS loss \cite{frietsch2008teaming}, UGV control with UAV's visual feedback \cite{xiao2017uav,xiao2015locomotive} using differential flatness \cite{rao2003visual}. However, UAVs in this category always hovered at a stationary and elevated viewpoint. Planning path to the viewpoint was trivial since the UAVs were also flying in open space without obstacles. 

In contrast to the aforementioned works, the aerial visual assistant in this work needs to navigate through unstructured or confined spaces with obstacles in order to provide visual assistance to the UGV operator from a series of good viewpoints. Instead of flying in wide open space or hovering at a stationary viewpoint, such motion entails taking risk, which may trade off good visual assistance. An understanding of both visual assistance viewpoint quality and motion risk in unstructured or confined environments is required to make rational decisions to enable risk-aware visual assistance. 

\subsection{Visual Assistance}
To our best knowledge, there is no existing formal theory of viewpoints for visual assistance and the majority (25 out of the 40 studies we reviewed) of robotic visual assistant implementations relied on ad hoc choices while the remainder (15 out of 40) relied on models of the work envelope. No robotic visual assistants considered psychophysical aspects in viewpoint selection. Four categories of robotic visual assistant implementations in the literature lacked principles to select ideal viewpoints: (1) Static visual assistants ~\cite{sato2019derivation,sato2019experimental,sato2017a,dima2019view} cannot adapt viewpoints to changing actions of the primary robot. (2) Manual visual assistants~\cite{shiroma2005cooperative,murphy2013apprehending,murphy2016affordances,murphy2016can,murphy2017within,perkins2013active,murphy2008cooperative,murphy2014disaster,leon2016from} left the choice of a viewpoint to humans who were shown to pick suboptimal viewpoints~\cite{mckee2003human}. (3) Reactive autonomous visual assistants~\cite{triggs1995automatic,hershberger2000distributed,simmons2001first,maeyama2016view,gawel2018aerial,yang2015performance,yang2015inducement,sato2016gaze,ji2018learning,abifarraj2016a,rakita2018an,nicolis2018occlusion} only tracked and zoomed on the action ignoring the question of what is the best viewpoint. (4) Deliberative autonomous visual assistants ~\cite{mckee1994visual,brooks1995toward,mckee1995visual,mckee1995human,brooks2001the,brooks2002visual,mckee2003human,rahnamaei2014automatic,ito2015optimal,saran2017viewpoint,samejima2016multi,samejima2018visual,rakita2019remote,thomason2017adaptive,thomason2019a} might be globally optimal in terms of geometry if the work envelope model is available but the existing studies did not consider other attributes, particularly psychophysical aspects when selecting viewpoints.

Cognitive science started to explore the psychophysical aspects of different views by categorizing a visual goal for an action as an affordance ~\cite{murphy2013apprehending,murphy2016affordances,murphy2016can,murphy2017within}. An affordance is a visual cue that allows to directly perceive the possibility of the action ~\cite{gibson2014ecological,murphy2019introduction}. T. Murphy conducted a human subject study to measure the tele-operator’s capability to correctly determine \emph{Reachability} affordance. Participants had difficulty determining \emph{Reachability} using onboard cameras. The performance improved when subjects were allowed to use a virtual external camera. However, this study only considered one affordance and did not explore which viewpoints were best for this affordance. This work explores affordances as the basis for the formal theory of viewpoints that can be generalized to actions comprising all tasks.

\subsection{Motion Risk}
Navigating in unstructured or confined environments inherently entails taking risk, which could be explicitly represented as a function of \emph{state} as in the literature. Those types of risk focused on collision with obstacles \cite{soltani2004fuzzy,de2011minimum} or with another vehicle \cite{pereira2013risk}. This is possible to happen in a particular \emph{state} of interest on the entire path. But when robots are locomoting in unstructured or confined environments, risk could stem from a variety of sources other than collision alone, e.g. getting stuck, sensor degradation, actuator malfunction, etc, and these may not only depend on the current \emph{state} the robot locates at. In our previous work \cite{xiao2019autonomous,xiao2019explicit2}, we expanded the list of risk elements, not being limited to only collision-related, and proposed the ideas of \emph{action}- and \emph{path}-dependent risk elements. However, risk of executing an entire path was still assumed to be additive with respect to risk at individual states, whose validity is still questionable. Expanding on these ideas and aiming at overcoming the ill-supported additivity assumption, this work further introduces a formal risk reasoning framework using propositional logic and probability theory. This framework is applicable for general mobile robot platforms and could be specifically used for our tethered aerial visual assistant. 

\subsection{Aerial Robots with Tether}
This work uses a tethered UAV as visual assistant, with the purpose of matching its battery duration with UGV's and as a failsafe in case of malfunction in mission-critical tasks. Although tethered UAVs have been studied in the literature, tether is either not incorporated into the locomotion at all, e.g., only for power \cite{zikou2015power,kiribayashi2018design} and safety \cite{pratt2008use} considerations, or included but only for stabilization \cite{lupashin2013stabilization} or steady flight in obstacle-free space \cite{schulz2015high}. The disadvantages brought by the tether were not deeply investigated, or even not looked into for UAVs at all. One important problem with tether is contact or entanglement with the environment. To the author's best knowledge, no motion planning and execution algorithms for autonomous flight of tethered UAV in indoor unstructured or confined environments exist in the current literature. In this work, our visual assistant aims at utilizing the advantages and mitigating the disadvantages brought by the tether, in terms of tether-based indoor localization \cite{xiao2018indoor}, motion primitives \cite{xiao2019benchmarking}, and environment contact planning \cite{xiao2018motion}.

\section{Co-Robots Team}
\label{sec::team}
\begin{figure}
\centering
\includegraphics[width=0.6\columnwidth]{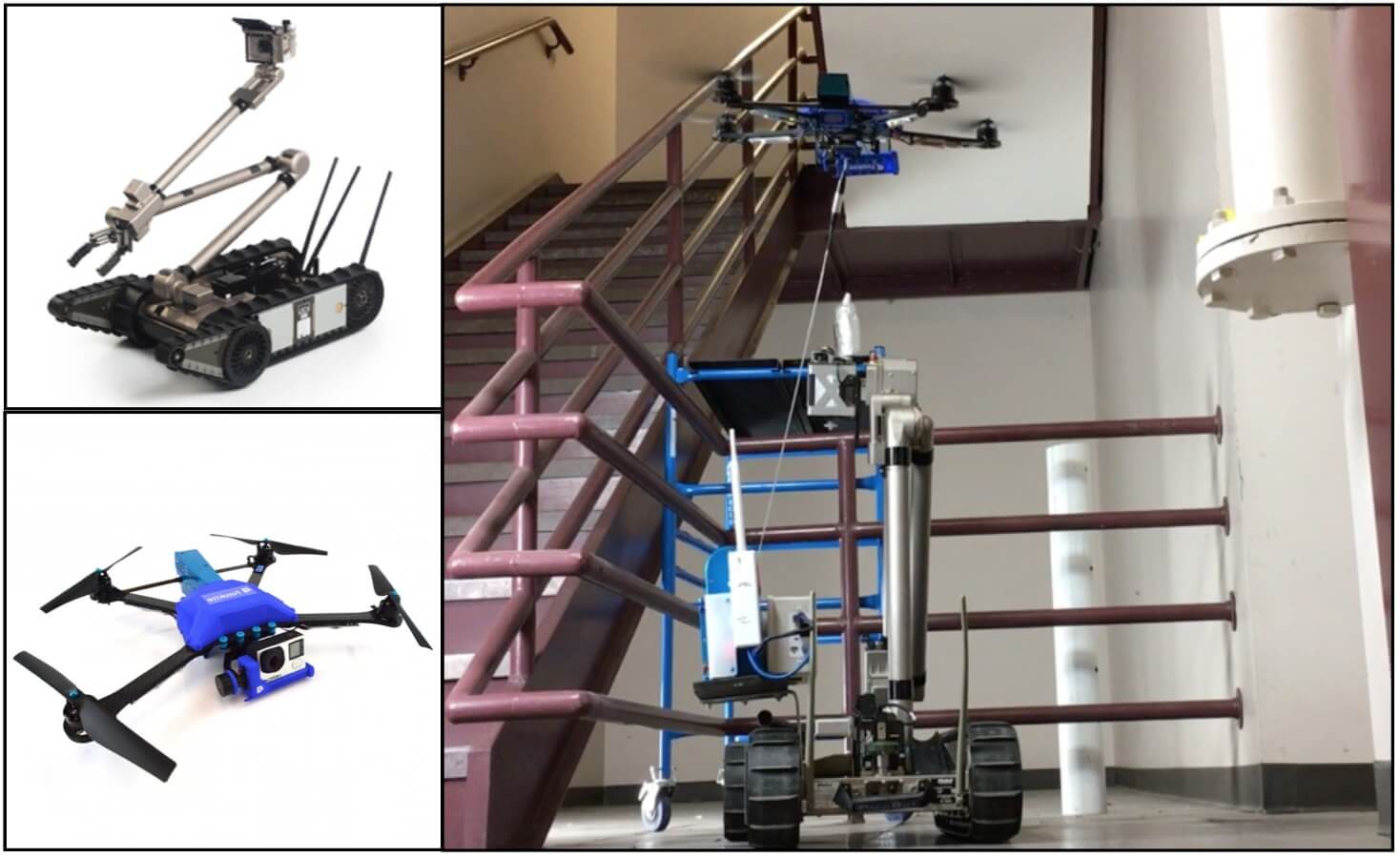}
\caption{The Co-Robots Team: Tele-operated primary robot, Endeavor PackBot 510 (upper left), and autonomous tethered aerial visual assistant, Fotokite Pro (lower left), picking up a sensor and dropping it into a radiation pipe in a confined staircase (right). }
\label{fig::team}
\end{figure}

This section presents the co-robots team (Fig. \ref{fig::team}): a tele-operated ground primary robot, an autonomous tethered aerial visual assistant, and a human operator of the primary robot under the visual assistance of the aerial vehicle, all of which are coordinated under an entire networked system architecture spanning from the remote field to the control center. 

\subsection{Tele-operated Ground Primary Robot}
In this co-robots team, a tele-operated Endeavor PackBot 510 (Fig. \ref{fig::team} upper left) is used as the primary robot. The chassis equipped with two main differential treads allows the ground robot to perform zero radius turns and achieve maximum speed up to 9.3km/h. In order to negotiate uneven terrain, or even stairs, two articulated and treaded flippers give the platform the ability to ascend and descend smoothly. To project human presence and physically act upon the remote environments, PackBot has a three-link manipulator on top of the chassis, with an articulated gripper on the second link and an onboard camera on the third. The manipulator has a lift capacity of 5kg at full extension and 20kg close-in. Motor encoders on the arm provide precise position of the articulated joints. Four onboard cameras provide first-person-views, but are all limited to the robot body. On the chassis, a Velodyne Puck LiDAR constantly scans the 3-D environments, providing the map for the co-robots team to navigate through. Four BB-2590 batteries provide up to 8 hrs run time. Other than Endeavor PackBot 510, other tele-operated ground robots, such as QinetiQ Talon, could also be used as primary robot in the co-robots team. 

\subsection{Autonomous Aerial Visual Assistant}
A tethered UAV, Fotokite Pro, serves as the autonomous aerial visual assistant (Fig. \ref{fig::team} lower left) for the tele-operated ground robot. The visual feed is streamed from the UAV's onboard camera actuated by a 2-DoF gimbal (pitch and roll). The camera's yaw is controlled dependently by the vehicular yaw. The tether provides constant wired power transmission and therefore matches the run time of the UAV with the UGV. Additionally, tether serves as a fail-safe in mission-critical environments. The tethered UAV could be deployed from a landing platform mounted on the ground robot's chassis. The UAV's flight controller is based on the tether sensory feedback relative to the origin on the landing platform, including the tether length, azimuth and elevation angles. 

\subsection{Human Operator}
\begin{figure}
\centering
\subfloat[PackBot uPoint Controller Interface]{\includegraphics[width=0.43\columnwidth]{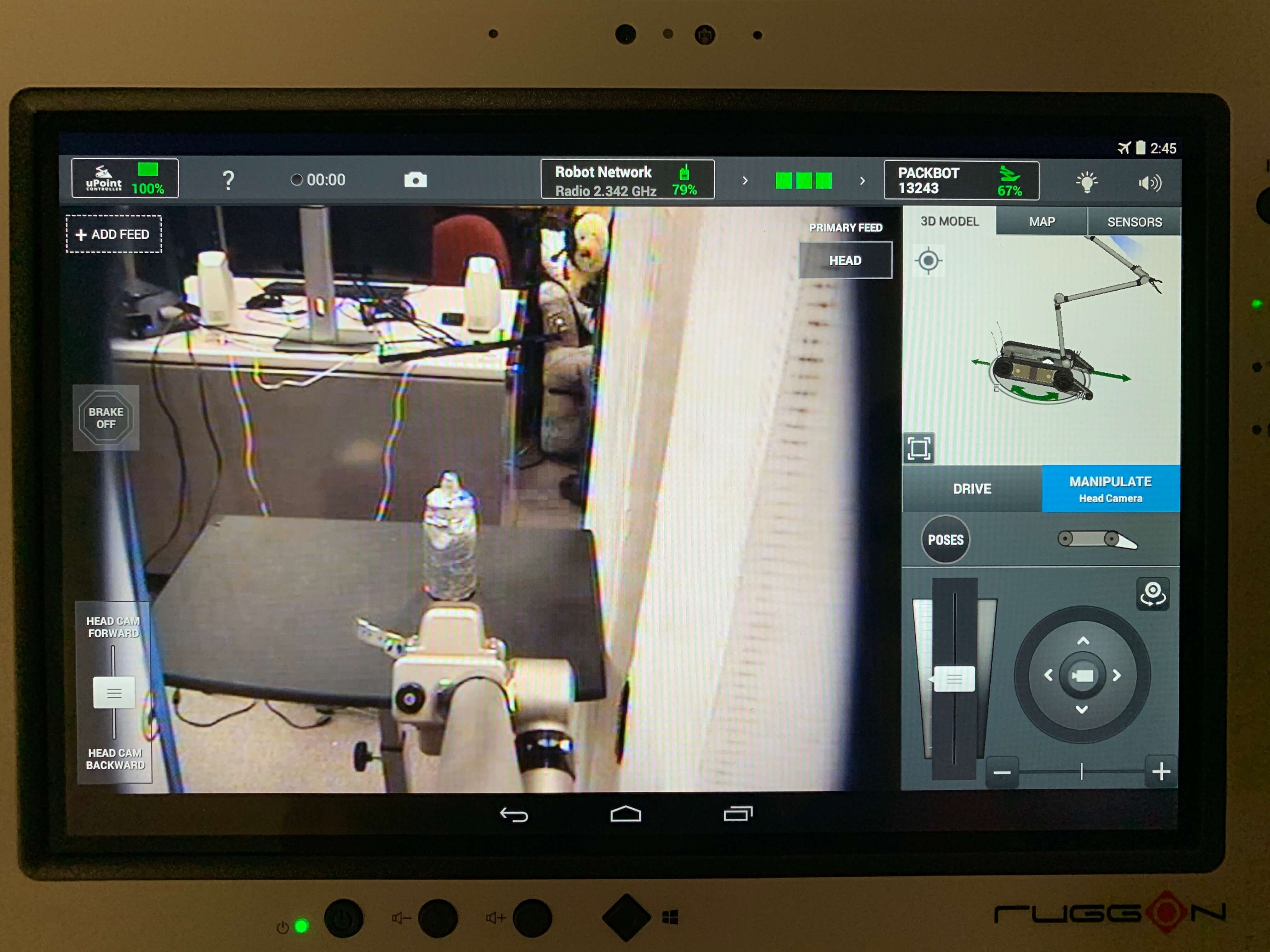}%
\label{fig::upoint}}
\subfloat[Visual Assistant Interface]{\includegraphics[width=0.57\columnwidth]{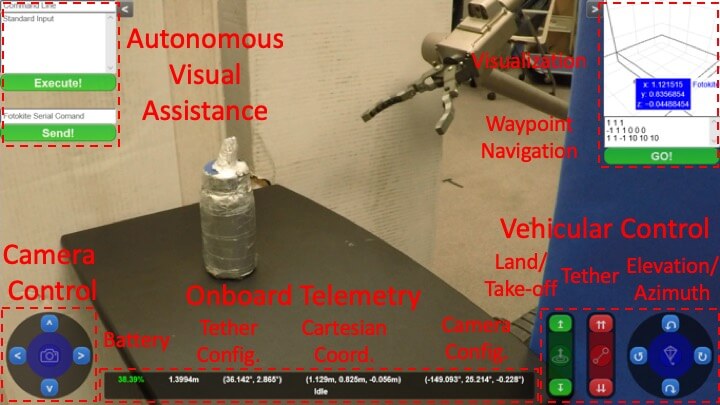}%
\label{fig::mickie}}
\caption{Interfaces with the Human Operator}
\label{fig::interfaces}
\end{figure}

The human operator tele-operates the primary ground robot with the visual assistance of the UAV. In addition to the default PackBot uPoint controller with onboard first-person-view (Fig. \ref{fig::upoint}), the visual feedback from the visual assistant's onboard camera is also available for enhanced situational awareness (Fig. \ref{fig::mickie}). For example, the visual assistant could move to a location perpendicular to the tele-operation action, providing extra depth perception to the operator (Fig. \ref{fig::interfaces}). The visual assistant could be either manually controlled or automated. For the focus of this research, autonomous visual assistance, a 3-D map is provided by the primary robot's LiDAR, and a risk-aware path is planned using a pre-established viewpoint quality map (discussed in the following sections).

\subsection{System Architecture}
\begin{figure}
   \centering
    \includegraphics[width=0.93\columnwidth]{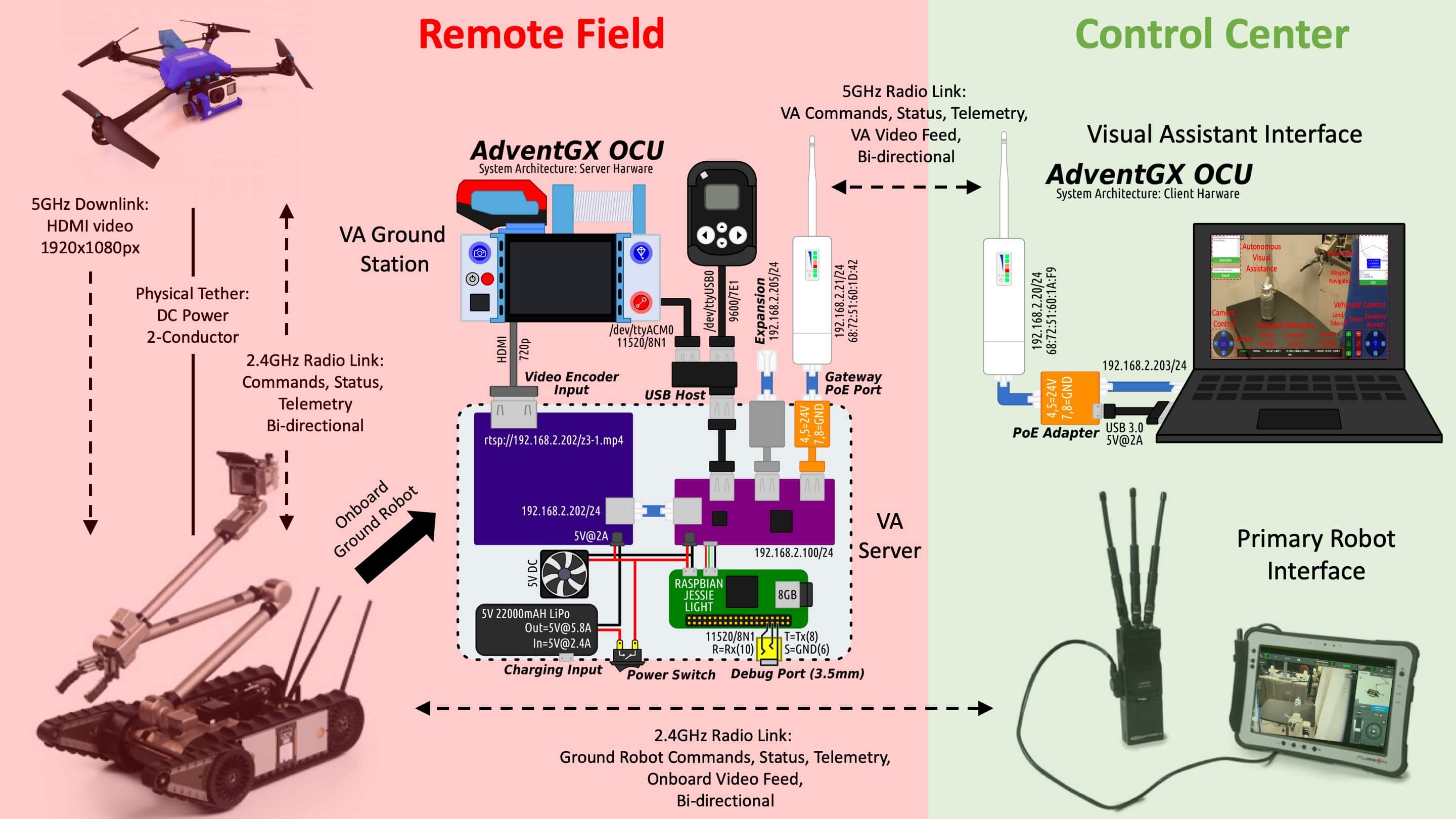}
    \caption{System Architecture}
    \label{fig::implementation_system_architecture}
\end{figure}

All the components of the co-robots team are connected based on the system architecture shown in Fig. \ref{fig::implementation_system_architecture}. The entire system locates in two separate locations, the remote field where the tele-operation mission takes place and the control center where the tele-operator is physically located. The communication between those two locations are through multiple bi-directional radio links. 

\subsubsection{In Remote Field}
The tethered aerial visual assistant's ground station (with tether reel) is mounted on the primary ground robot. The UAV is powered via the physical tether, while onboard commands, status, and telemetry are transmitted by 2.4GHz bi-directional radio link. The HDMI video from the visual assistant's camera is streamed via 5GHz radio downlink to the ground station. The primary robot, PackBot, is tele-operated by 2.4GHz bi-directional radio link for controls and video. 

A server built in collaboration with AdventGX is connected with the UAV's ground station via HDMI and USB cables, for video encoding and UAV control (via Fotokite SDK), respectively.  The black RadEye SPRD spectroscopic radiation detector is not used in this particular work. The server is composed of a video encoder (shown in purple in Fig. \ref{fig::implementation_system_architecture}) and a Raspberry Pi computer (shown in green in Fig. \ref{fig::implementation_system_architecture}). The video encoder encodes the HDMI video output from the ground station so it could be transmitted wirelessly to the visual assistant's Operator Control Unit (OCU). The Raspberry Pi computer connects to the ground station via USB and uses Fotokite SDK to receive sensor data and send control commands. All video and telemetry data are ported to an antenna and transmitted to the visual assistant's OCU wirelessly via a 5GHz radio link. The link is bi-directional, whose other direction is used for transmitting control commands from the OCU to the server. The server is self-powered with its own battery and cooled by a 5V DC fan. 

All the executables of the UAV controls are compiled from C++ code and then stored on the server's Raspberry Pi computer. Direct tether, vehicle and camera commands (tele-operation) are ported directly from the OCU to the server and then to the Fotokite ground station. Autonomous flight, including landing, taking-off, autonomous waypoint navigation and visual assistance, is triggered by the OCU, and then the corresponding executable on the server is called. 

\subsubsection{In Control Center}
In the control center, the tele-operator uses the uPoint interface to control the primary ground robot via 2.4GHz bi-directional radio link. The current system has a separate OCU for the visual assistant, which is planned to be integrated with the ground robot interface in the future. The visual assistant OCU is a laptop connected with an antenna for 5GHz bi-directional radio link. The interface is implemented on a specific web socket and could be displayed via web browser. As mentioned earlier, the interface allows both tele-operation and autonomous navigation. Paths are pre-planned based on a given 3D map, saved in the OCU, and uploaded to the server. The autonomous navigation executable on the server called by the OCU command takes the path as input argument. The tele-operator only needs to select the autonomous navigation and all the following process for the visual assistant is carried out autonomously.

\section{Viewpoint Quality}
\label{sec::viewpoint_quality}
In order to establish a formal theory of tele-operation visual assistance viewpoint quality based on the cognitive science concept of Gibsonian affordances, a formal study is conducted with 35 human subjects for 4 common affordances and viewpoints are clustered into manifolds based on their merit for visual assistance using agglomerative hierarchical clustering.

\subsection{Human Study}
\begin{figure}
\centering
\subfloat[Reachability]{\includegraphics[width=0.45\textwidth]{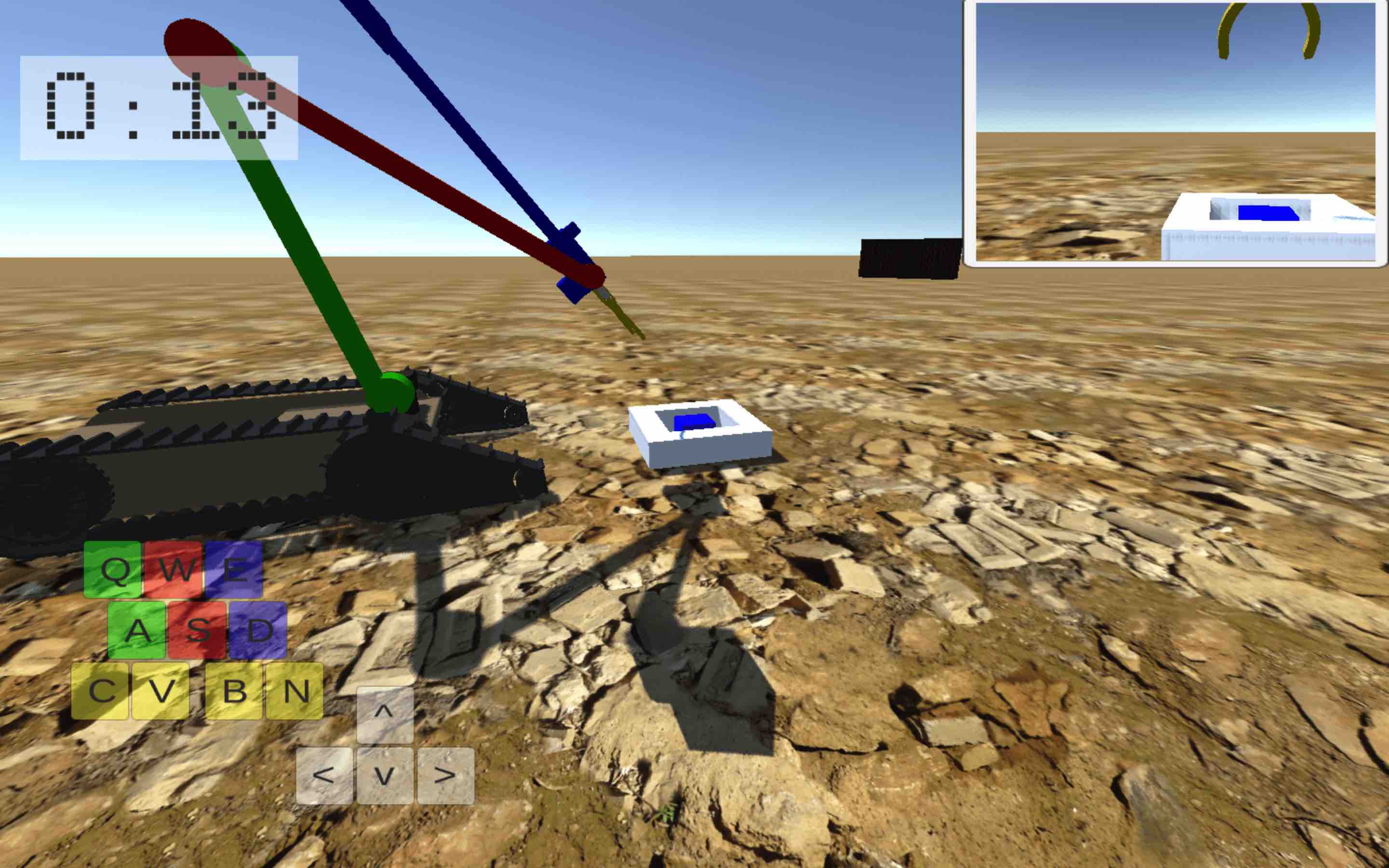}%
\label{fig::reachability}}
\hspace{1pt}
\subfloat[Passability]{\includegraphics[width=0.45\textwidth]{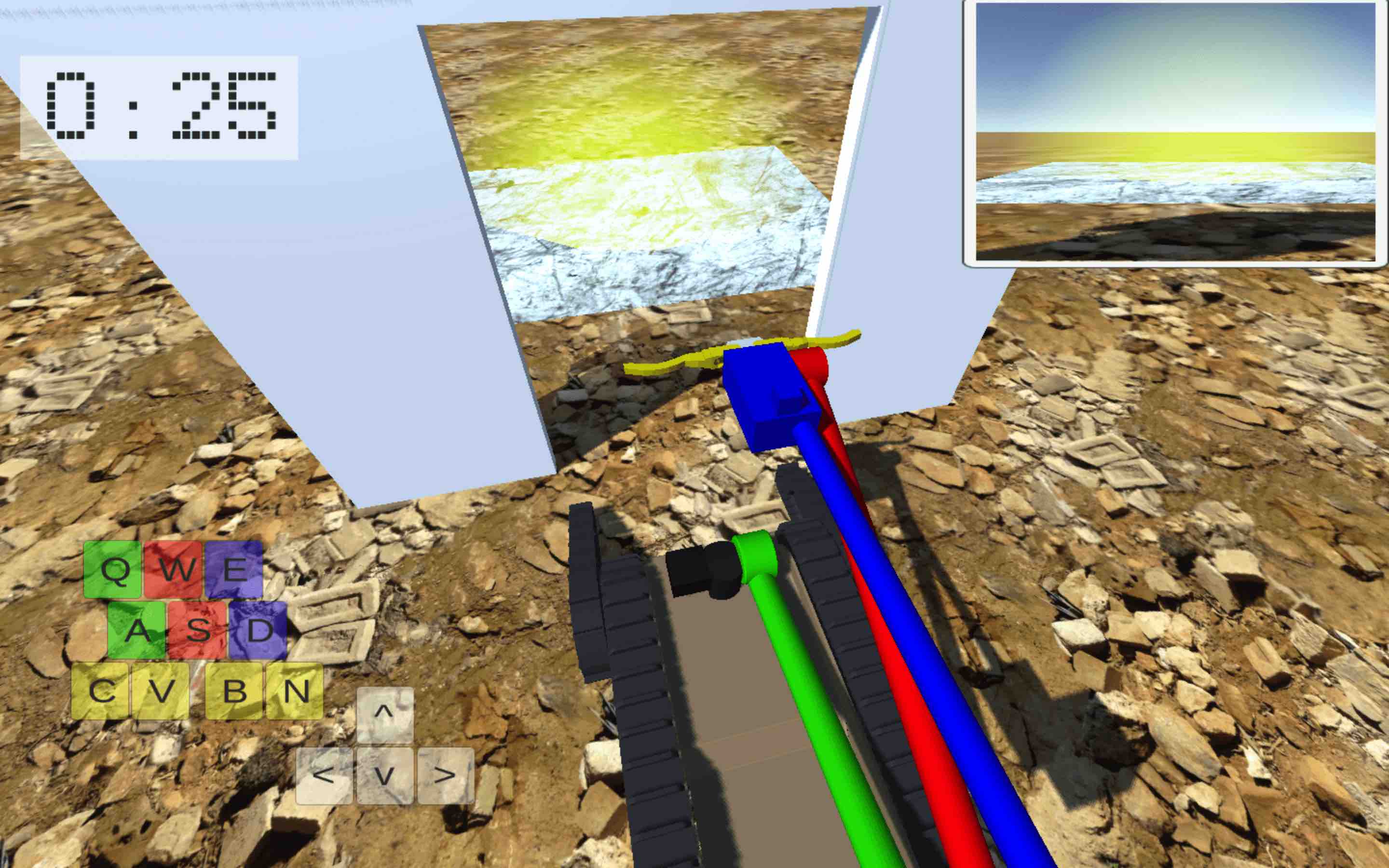}%
\label{fig::passability}}\\
\subfloat[Manipulability]{\includegraphics[width=0.45\textwidth]{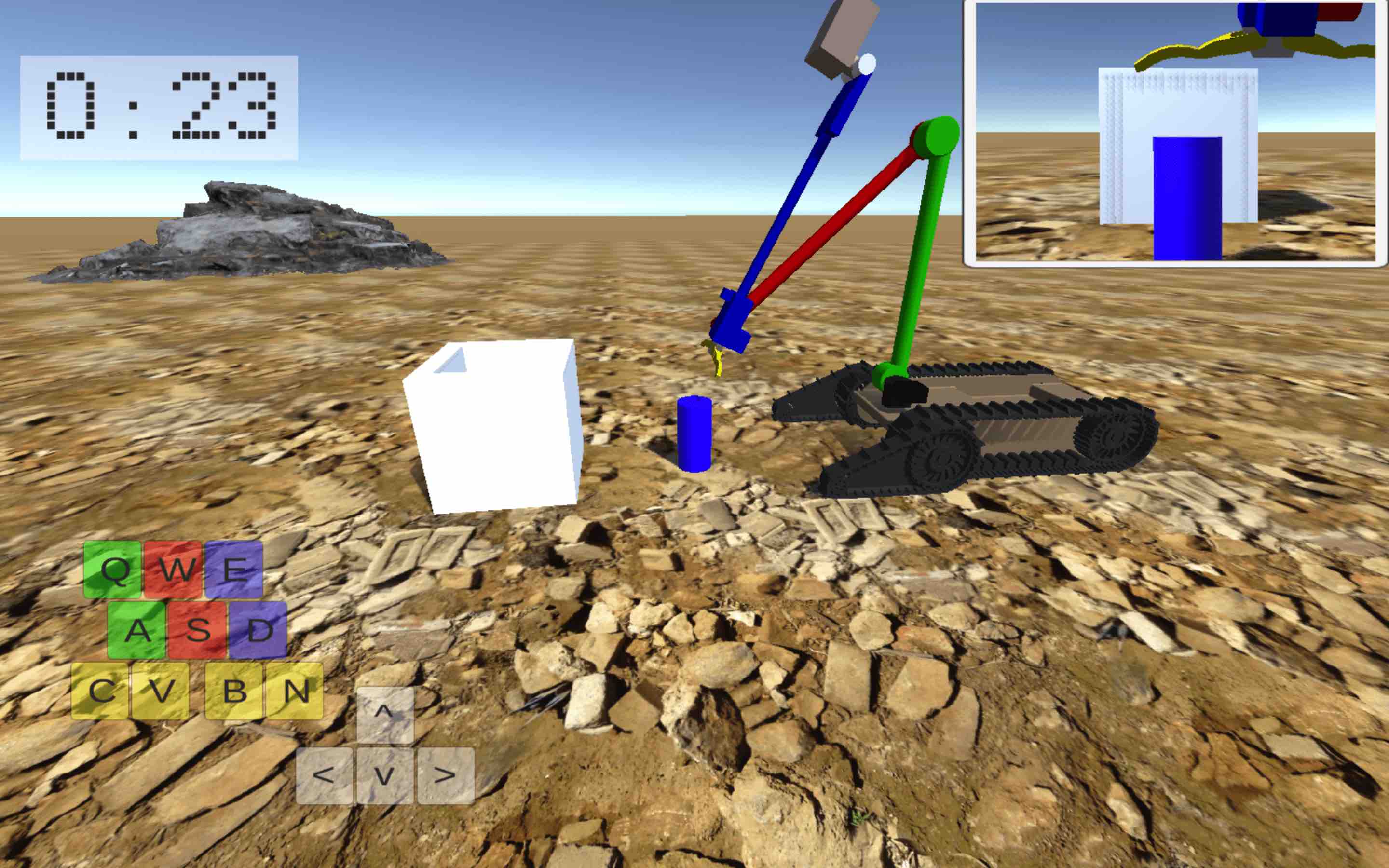}%
\label{fig::manipulability}}
\hspace{1pt}
\subfloat[Traversability]{\includegraphics[width=0.45\textwidth]{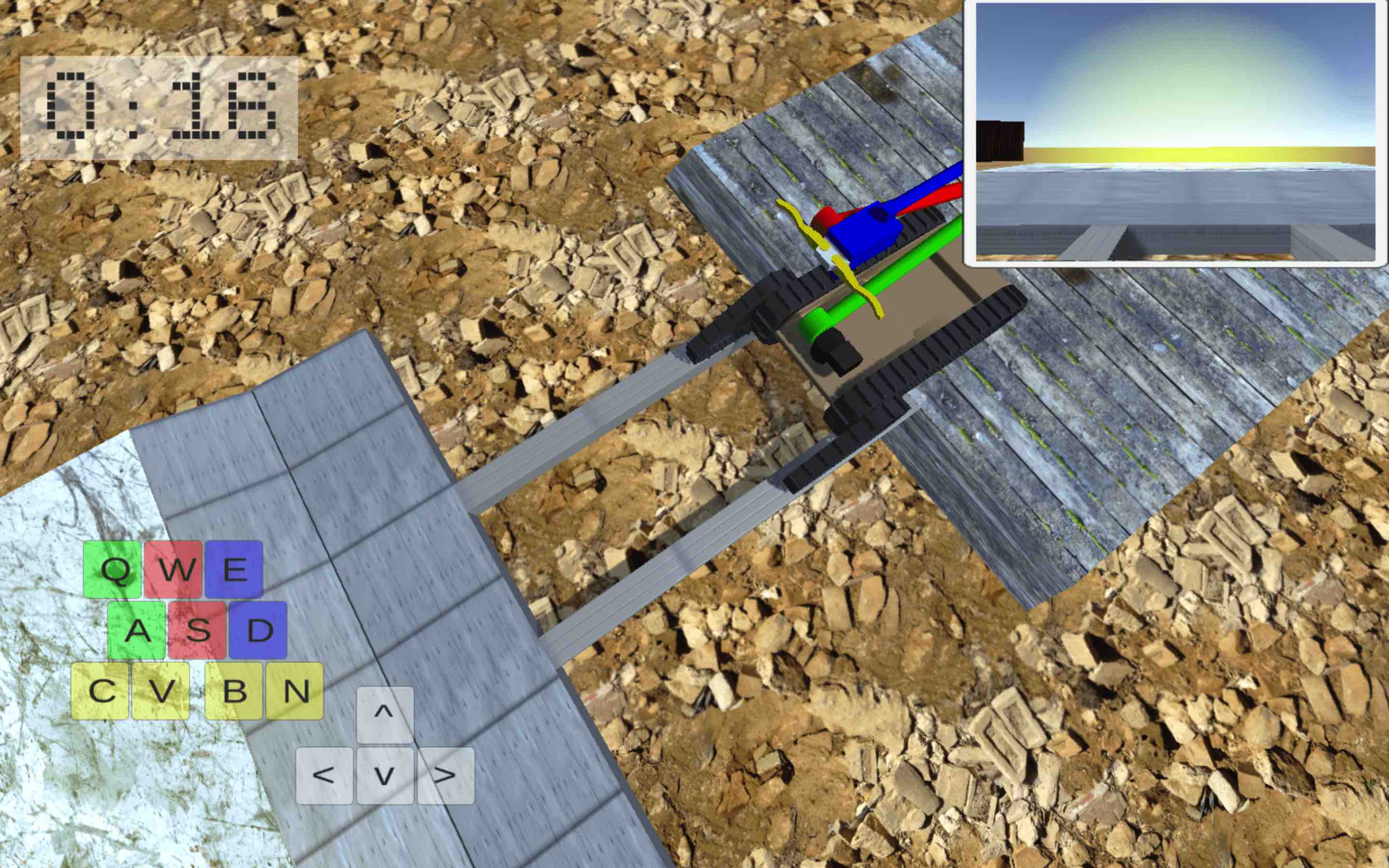}%
\label{fig::traversability}}
\caption{Four Tasks Corresponding to Four Gibsonian Affordances}
\label{fig:affordances}
\end{figure}

A human study with 35 subjects is conducted with a goal to sufficiently sample human performance for 30 viewpoints $v$ to formalize the value of viewpoints $|v|$ for 4 common affordances $a$ so that spatial clusters (manifolds) can be learned. Collected data is filtered based on its validity and 23 out of 35 subjects' data are used for data analysis, which is sufficient for the minimum required subject number, 18, from power analysis. Expert robot operators perform four types of tasks, each associated with one of the four affordances (Figure \ref{fig:affordances}) using computer-based simulation implemented in Unity and running on Amazon Web Services. For \emph{Reachability}, the task is to touch a cube using the gripper without hitting the neighboring blocks. For \emph{Passability}, the task is to pass through an opening without hitting the walls. For \emph{Manipulability}, the task is to pick up a cylinder and drop it in a bin without hitting the bin with the gripper. For \emph{Traversability}, the task is to cross a ridge without falling on the ground. 

The independent variable is the position of the external viewpoint provided to the subject. A total of 30 possible viewpoints, $v_i$, are equidistantly dispersed on a hemisphere with a fixed radius of $r =$~\SI{1.5}{\meter} centered in the task location. Those 30 viewpoints are divided into 5 groups (6 viewpoints per group) based on their relative position to the task location: left, right, front, back, and top. Each subject performs each of the 4 tasks from each of the 5 viewpoint groups (20 rounds total). A particular viewpoint from each group is always selected randomly. The order of the tasks and viewpoint groups is randomized for each subject to reduce the order effect.

There are two dependent variables both indicating the subject's performance: the time to complete the task and the number of errors. What is considered as an error depends on the affordance: For \emph{Reachability} and \emph{Manipulability}, the number of errors is the number of manipulator collisions. For \emph{Passability}, the number of errors is the number of robot collisions. For \emph{Traversability}, the number of errors is the number of falls of the robot. 

The metric quantifying the value of a viewpoint is the subjects' performance defined as a weighted sum of normalized time and normalized number of errors:

\begin{equation}
\label{eq:performance}
\tensor*[_j]{P}{^a_i} = 0.4 \left(\frac{\tensor*[_j]{t}{^a_i} - \mean_{a^\prime,i^\prime}{\tensor*[_j]{t}{^{a^\prime}_{i^\prime}}}}{\std_{a^\prime,i^\prime}{\tensor*[_j]{t}{^{a^\prime}_{i^\prime}}}}\right) + 0.6 \left(\frac{\tensor*[_j]{e}{^a_i} - \mean_{a^\prime,i^\prime}{\tensor*[_j]{e}{^{a^\prime}_{i^\prime}}}}{\std_{a^\prime,i^\prime}{\tensor*[_j]{e}{^{a^\prime}_{i^\prime}}}}\right),
\end{equation}

where $j$ is subject index, $i$ is viewpoint index, $a$ is affordance index, $\tensor*[_j]{P}{^a_i}$ denotes the performance of subject $j$ for affordance $a$ from viewpoint $v_i$, $\tensor*[_j]{t}{^a_i}$ is the time subject $j$ takes to complete the task associated with affordance $a$ from viewpoint $v_i$, and $\tensor*[_j]{e}{^a_i}$ is the number of errors subject $j$ makes when performing the task associated with affordance $a$ from viewpoint $v_i$. $\mean$ and $\std$ are for normalization and denote the mean and standard deviation values of this particular subject $j$'s performance on all affordances from all viewpoints. Errors are weighted slightly higher than completion time to penalize completing a task faster at the expense of more errors.

The study results in 460 data points where one data point is a tuple $(a, v_i, j, \tensor*[_j]{P}{^a_i})$ representing a score, $\tensor*[_j]{P}{^a_i}$, corresponding to the performance of subject $j$ at viewpoint $v_i$ for affordance $a$. A data point is rejected as an outlier if the corresponding  $\tensor*[_j]{t}{^a_i}$ value is more than three scaled median absolute deviations (MAD) away from its median. An example of an outlier would be if the subject gets distracted during task completion causing the performance to be significantly worse than it should be. Grouping the data results in a list of 30 viewpoints $v_i$ with associated $\abs{v_i^a} = \mean_{j}{\tensor*[_j]{P}{^a_i}}$ for each affordance $a$.


\subsection{Viewpoints Clustering}

Agglomerative hierarchical cluster analysis~\cite{bridges1966hierarchical} with average linkages~\cite{szekely2005hierarchical} is used to generate manifolds. This method is appropriate because it considers the proximity of viewpoints, does not require the number of clusters in advance, and handles outliers in a favorable way.

The input to the hierarchical cluster analysis is the 30 sample points for each affordance $a$ resulting from the human subject study, where one sample point is a four dimensional vector

\begin{equation}
s_i = (\overrightarrow{\bf{v_i}}, \abs{v_i^a}),~ i = 1, ..., 30,
\end{equation}

where $\overrightarrow{\bf{v_i}}$ is the $i$-th viewpoint position represented in spherical coordinates as $(r_i, \theta_i, \varphi_i)$, $\abs{v_i^a} = \mean_{j}{\tensor*[_j]{P}{^a_i}}$ is the value of the $i$-th viewpoint, and $\tensor*[_j]{P}{^a_i}$ is the performance of subject $j$ for affordance $a$ from viewpoint $v_i$.

Pairwise dissimilarity between all 30 sample points $s_i$ is computed using standardized Euclidean distance resulting in 435 dissimilarities

\begin{equation}
d_{{s_i}{s_j}} = \sqrt{(s_i - s_j) K^{-1} (s_i - s_j)^\prime},
\end{equation}

where $K$ is a 4-by-4 diagonal covariance matrix whose $n$-th diagonal element is $Q^2(n)$, standard deviation of the $n$-th dimension across the 30 sample points. Standardized Euclidean distance is used because the dimensions have different scales.

The samples are grouped into a binary hierarchical cluster tree using Unweighted Pair Group Method with Arithmetic mean (UPGMA) linkage. The average linkage between clusters $M_k$ and $M_l$ is defined as

\begin{equation}
d_{{M_k}{M_l}} = \frac{1}{\abs{M_k}\abs{M_l}} \sum_{s_i \in M_k} \sum_{s_j \in M_l} d_{{s_i}{s_j}}.
\end{equation}

\begin{figure} [htbp]
    \centering
    \includegraphics[width=\textwidth]{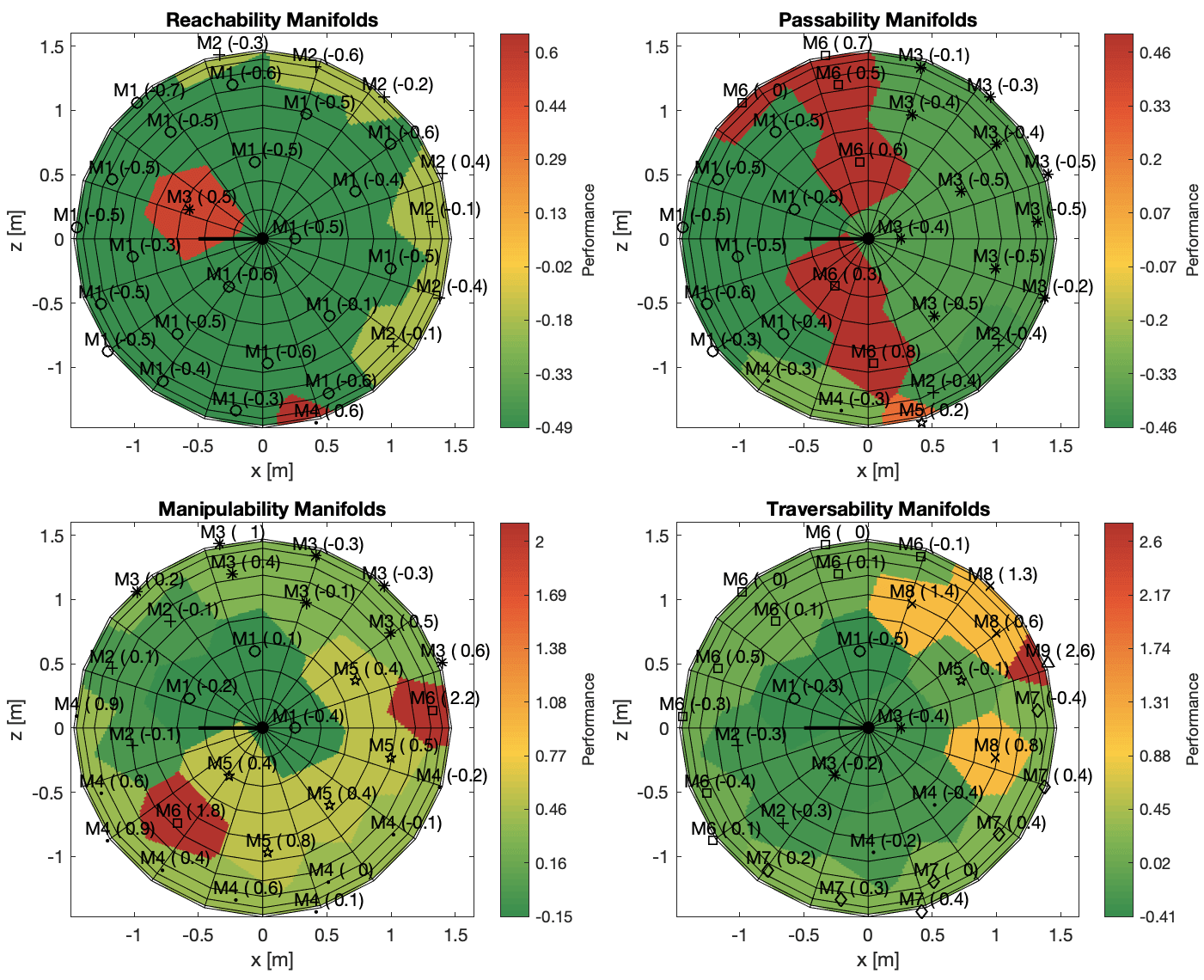}
    \caption{Resulting Manifolds for Each Affordance: Each graph shows a top-down view of the hemisphere of viewpoints around the task. The task is represented by the solid black dot centered in the hemisphere and its orientation is represented by the black bold solid line radiating from the solid black dot. Each of the 30 viewpoints for each affordance is represented by a black marker whose shape indicates which manifold the viewpoint belongs to (i.e., viewpoints belonging to the same manifold have the same marker). Each viewpoint's label is composed of a manifold number prefixed by capital M followed by the viewpoint's performance value $|v|$ in parentheses. The partitioning of the space into manifolds is visualized via a color map where a continuous area with the same color represents a single manifold and the color represents the manifold's value $M$ (mean of $|v|$ of viewpoints inside that manifold). Higher value means worse viewpoint quality. The manifolds are numbered based on ascending performance.}
    \label{fig:manifolds}
\end{figure}

The hierarchical cluster tree (dendrogram) is cut based on an inconsistency coefficient to find the borders of natural cluster divisions in the data set resulting in $N_m$ final clusters or manifolds, $M_i$, $i = 1, ..., N_m$, for each affordance $a$. The value of a manifold $M_i$ denoted by $|M_i|$ is defined as a mean of $|v_j|$ where $v_j$ are viewpoints inside the manifold. The resulting manifolds are shown in Figure~\ref{fig:manifolds}. Note that lower performance score (less completion time and fewer operation errors) means better viewpoint quality. 

For \emph{Reachability}, there are $N_m = 4$ manifolds with an inconsistency coefficient of $1.15$. The best manifold $M_1$ has the value of $-0.49$ and occupies the front and top part of the hemisphere. Slightly worse manifold $M_2$ has value $-0.19$ occupying the lower back hemisphere. The worst manifolds $M_3$ and $M_4$ have values $0.49$ and $0.6$ respectively forming isolated clusters.

For \emph{Passability}, there are $N_m = 6$ manifolds with inconsistency coefficient of $1.15$. Manifolds $M_1$, $M_2$, $M_3$, and $M_4$ all have good performance of $-0.46$, $-0.42$, $-0.38$, and $-0.3$ respectively occupying front and back portion of the hemisphere. Manifolds $M_5$ and $M_6$ have very bad performance of $0.21$ and $0.46$ respectively forming a narrow band across the hemisphere perpendicular to the task orientation.

For \emph{Manipulability}, there are 6 manifolds with an inconsistency coefficient of $1.15$. The best manifolds $M_1$ and $M_2$ have values of $-0.15$ and $-0.04$ occupying the top right portion of the hemisphere. The lower portions of the hemisphere are surrounded by slightly worse manifolds $M_3$ and $M_4$ with values $0.25$ and $0.36$. A worse still manifold $M_5$ with the value of $0.49$ occupies the top left and back portion of the hemisphere. The worst manifold is $M_6$ with value $2.00$ forming two disjoint regions.

For \emph{Traversability}, there are $N_m = 9$ manifolds with an inconsistency coefficient of $1.14$. The best manifolds $M_1$, $M_2$, $M_3$, $M_4$, and $M_5$ all have very similar values of $-0.41$, $-0.32$, $-0.31$, $-0.27$, and $-0.07$ and occupy the top portion of the hemisphere. Slightly worse manifolds $M_6$ and $M_7$ with values $0.01$ and $0.18$ encircle the lower portion of the hemisphere on the front, left and right. Bad manifolds $M_8$ and $M_9$ with values $1.04$ and $2.6$ occupy the lower back part of the hemisphere.

\section{Risk-Awareness}
\label{sec::risk-awareness}
In order to conduct trust-worthy visual assistance in unstructured or confined environments, the co-robots team adopts a formal risk reasoning framework \cite{xiao2019robot} and tailored the risk universe to form a subset of relevant risk elements for the tethered aerial visual assistant. The general risk reasoning framework is firstly recapitulated here, before the relevant subset of risk elements are formed for the tethered UAV. 

\subsection{Formal Risk Reasoning}

Risk of executing a path in unstructured or confined environments is formally defined as \emph{the probability of the robot not being able to finish the path} \cite{xiao2019robot}. Assuming the robot workspace is based on tessellation of the Cartesian space, a feasible path is composed of an ordered sequence of viable tessellations, called states and denoted as $s_i$: $P = \{s_0, s_1, ..., s_n\}$. In order to finish the path of $n$ states, the robot faces $r$ different risk elements at each state, which will possibly cause not finishing the path. Here, three types of events are defined with propositional logic: 
\begin{itemize}
\item $F$ -- the event where the robot finishes path $P$
\item $F_i$ -- the event where the robot finishes state $i$
\item $F_i^k$ -- the event where risk $k$ does not cause a failure at state $i$
\end{itemize}

The reasoning about motion risk is based on three assumptions, which are expressed by propositional logic: 
\begin{enumerate}
	\item Path is finished only when all states are finished: 
	$F = F_n \cap F_{n-1} \cap ... \cap F_1 \cap F_0$
	
	\item A state is finished only when all risk elements do not cause failure: 
	$F_i = F_i^1 \cap F_i^2 \cap ... \cap F_i^{r-1} \cap F_i^r$
	
	\item Finish or fail a state because of one risk element is conditionally independent of finish or fail that state because of any other risk element, given the history leading to the state:  
	$(F_i^1 \vert \bigcap \limits_{j=0}^{i-1} F_j) \perp \!\!\! \perp (F_i^2 \vert \bigcap \limits_{j=0}^{i-1} F_j) \perp \!\!\! \perp ... \perp \!\!\! \perp (F_i^r \vert \bigcap \limits_{j=0}^{i-1} F_j)$
	
\end{enumerate}

Before reasoning about risk, the probability of the robot \emph{being} able to finish the path $P(F)$ is firstly reasoned: 

\begin{equation}
\begin{split}
P(F) 
&= P(F_n\cap F_{n-1} \cap ... \cap F_0) \\
&= P(F_n \vert F_{n-1} \cap ... \cap F_0) \cdot ... \cdot P(F_1 \vert F_0) \cdot P(F_0)\\
&= \prod_{i=0}^{n} P(F_i \vert \bigcap_{j=0}^{i-1} F_j) \\
& = \prod_{i=0}^{n} P(F_i^1 \cap F_i^2 \cap ... \cap F_i^r \vert \bigcap_{j=0}^{i-1} F_j) \\
& = \prod_{i=0}^{n} P(F_i^1 \vert \bigcap_{j=0}^{i-1} F_j) \cdot ... \cdot P(F_i^r \vert \bigcap_{j=0}^{i-1} F_j) \\
& = \prod_{i=0}^{n} \prod_{k=1}^{r} P(F_i^k\vert \bigcap_{j=0}^{i-1} F_j)
\end{split}
\end{equation}

The first, second, fourth, and fifth equal signs are due to assumption 1, probability chain rule, assumption 2, and assumption 3, respectively. Therefore, the formal risk definition, the probability of \emph{not} being able to finish the path, is the probabilistic complement: 

\begin{equation}
\begin{split}
P(\bar{F}) 
&= 1 - P(F) \\
&= 1 - \prod_{i=0}^{n} \prod_{k=1}^{r} P(F_i^k \vert \bigcap_{j=0}^{i-1} F_j) \\
&= 1 - \prod_{i=0}^{n} \prod_{k=1}^{r} (1-P(\bar{F_i^k} \vert \bigcap_{j=0}^{i-1} F_j))
\end{split}
\label{eqn::pfbar}
\end{equation}

In terms of risk representation, the risk of path $P$ is denoted as $risk(P)$ and is equal to $P(\bar{F})$. $P(\bar{F_i^k} \vert \bigcap\limits_{j=0}^{i-1} F_j)$ means the probability of risk $k$ causes failure at state $i$, given the history of finishing $s_0$ to $s_{i-1}$. It is therefore denoted as the $k$-th risk robot faces at state $i$ given that $s_0$ to $s_{i-1}$ were finished: $r_k(\{s_0, s_1, ..., s_i\})$. Writing in risk representation form will yield: 

\begin{equation}
risk(P) = 1 - \prod_{i=0}^{n} \prod_{k=1}^{r} (1-r_k(\{s_0, s_1, ..., s_i\}))
\label{eqn::risk_representation}
\end{equation}

\begin{figure}
\centering
	\includegraphics[width = 0.7 \columnwidth]{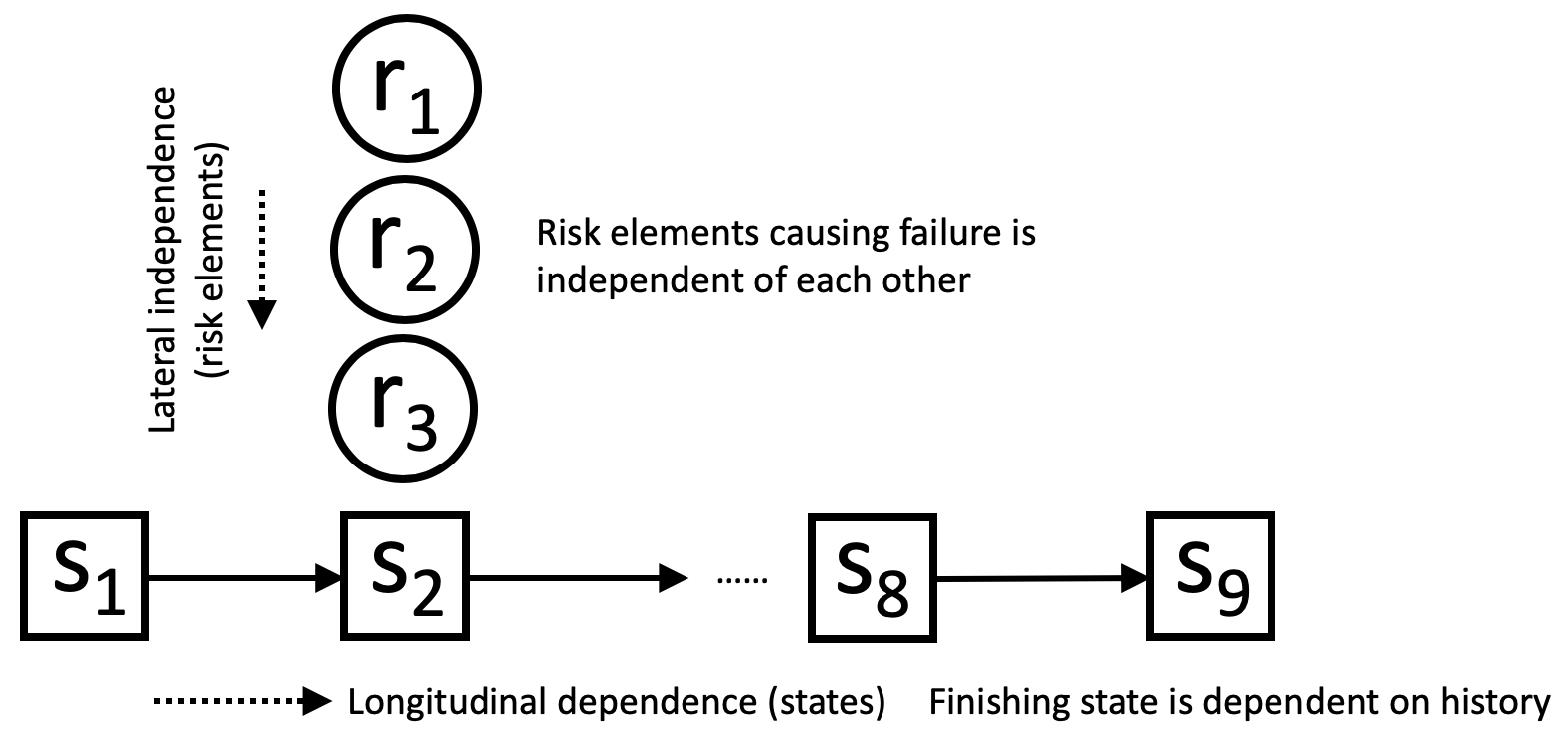}
	\caption{Longitudinal Dependence on History States and Lateral Independence among Risk Elements}
	\label{fig::dependencies}
\end{figure}

A graphical illustration of the (in)dependencies is shown in Fig. \ref{fig::dependencies}. Based on different depths of longitudinal dependence, it is practical to subdivide the risk universe into three categories:

\begin{itemize}
\item \emph{Locale-dependent Risk Elements}: $P(\bar{F_i^k} \vert \bigcap_{j=0}^{i-1} F_j) = P(\bar{F_i^k})$
\item \emph{Action-dependent Risk Elements}: $P(\bar{F_i^k} \vert \bigcap_{j=0}^{i-1} F_j) = P(\bar{F_i^k} \vert F_{i-2} \cap F_{i-1})$
\item \emph{Traverse-dependent Risk Elements}: $P(\bar{F_i^k} \vert \bigcap_{j=0}^{i-1} F_j) = P(\bar{F_i^k} \vert F_{i-1} \cap F_{i-2} \cap ... \cap F_1 \cap F_0) $
\end{itemize}

which is dependent on the current state alone, two states back, and the entire history, respectively. 

\subsection{Relevant Risk Elements}
\begin{figure}
\centering
\includegraphics[width=1\columnwidth]{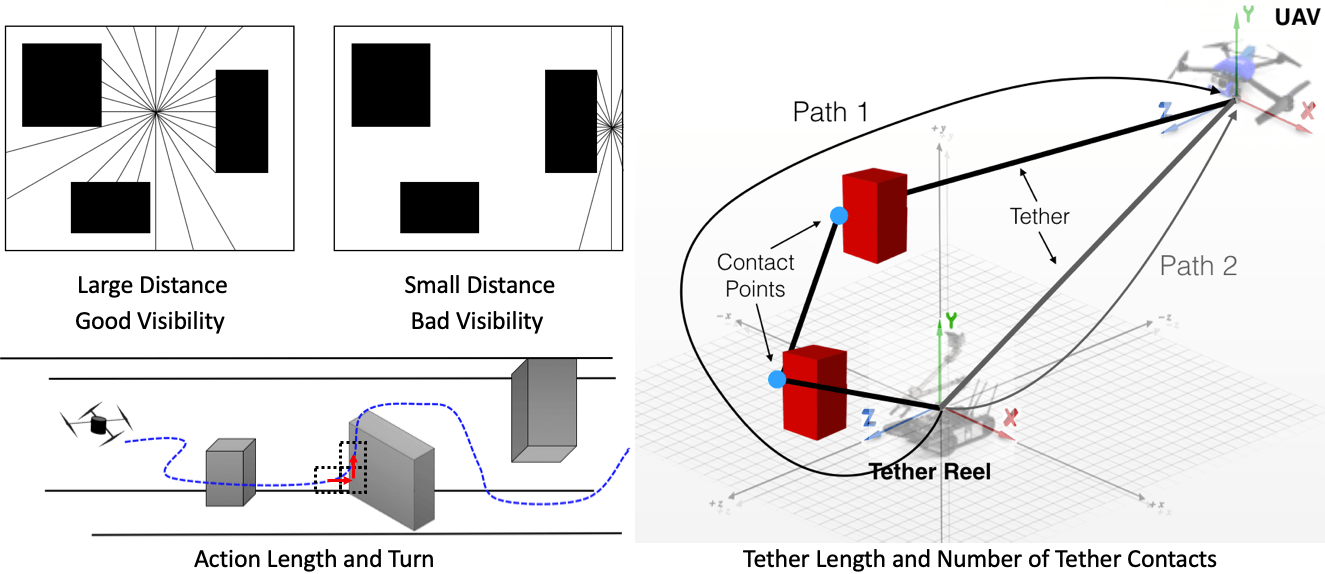}
\caption{Relevant Risk Elements: Distance to closest obstacles and visibility (upper left), action length and turn (lower left), tether length and number of tether contacts (right)}
\label{fig::risk_elements}
\end{figure}

Although there exists a non-exclusive universe of risk elements for general robotic applications \cite{xiao2019robot}, for the tethered aerial visual assistant addressed in this work, each of the three risk categories has two relevant risk elements. They are distance to closest obstacle and visibility (Fig. \ref{fig::risk_elements} upper left, distance computed from shortest and visibility from average length of isovists lines) for locale-dependence, action length and turn (Fig. \ref{fig::risk_elements} lower left, length computed from length of red action arrows and turn from two pairs of consecutive states) for action-dependence, tether length and number of tether contacts (Fig. \ref{fig::risk_elements} right, both computed from the entire path leading to the state) \cite{xiao2018motion} for traverse-dependence. 

\subsection{Risk-Aware Planning}
Two characteristics of the risk reasoning framework make planning with absolute minimum risk difficult: non-additivity and history-dependency. It is computationally intractable to find minimum-risk path for the general full-history-dependent risk elements. Trade-off between computational complexity and history dependency depth exists, therefore only minimum-risk path with respect to a limited history-dependency depth could be practically found within reasonable amount of time. Approach to expand planner's history dependency into action-dependent risk elements has been proposed by augmenting single state into multiple directional components, with potential methods to further deepen risk's history dependency at the cost of computation \cite{xiao2019robot}. 

Given a viewpoint quality map as reward and motion execution risk as a function of the entire path, the visual assistant plans minimum-risk path to each state \cite{xiao2019robot}, evaluates the reward collected \cite{xiao2019explicit}, and then picks the one with optimal utility value, defined as the ratio between reward and risk. Executing the optimal utility path approximates the optimal visual assistance behavior. 

\section{Tethered Motion}
\label{sec::tethererd_motion}
The aerial visual assistant operates under a taut tether. The benefits brought by the tether is three-fold: (1) The tether provides wired power delivery from the UGV to the UAV, allowing the co-robots team to work together in the remote field for an extended period of time without recharge, thanks to the large capacity battery onboard the UGV. (2) In case of malfunction, such as a UAV crash, tether could be used to retrieve the UAV. (3) As discussed below, the tether is utilized as a means of low-cost localization in GPS-denied environments. This section describes a low level motion suite used by the visual assistant to enable tethered flight in unstructured or confined environments without reliable GPS. 

\subsection{Tether-Based Indoor Localization}
\begin{figure}
\centering
\subfloat[Localization Model]{\includegraphics[width=0.33\columnwidth]{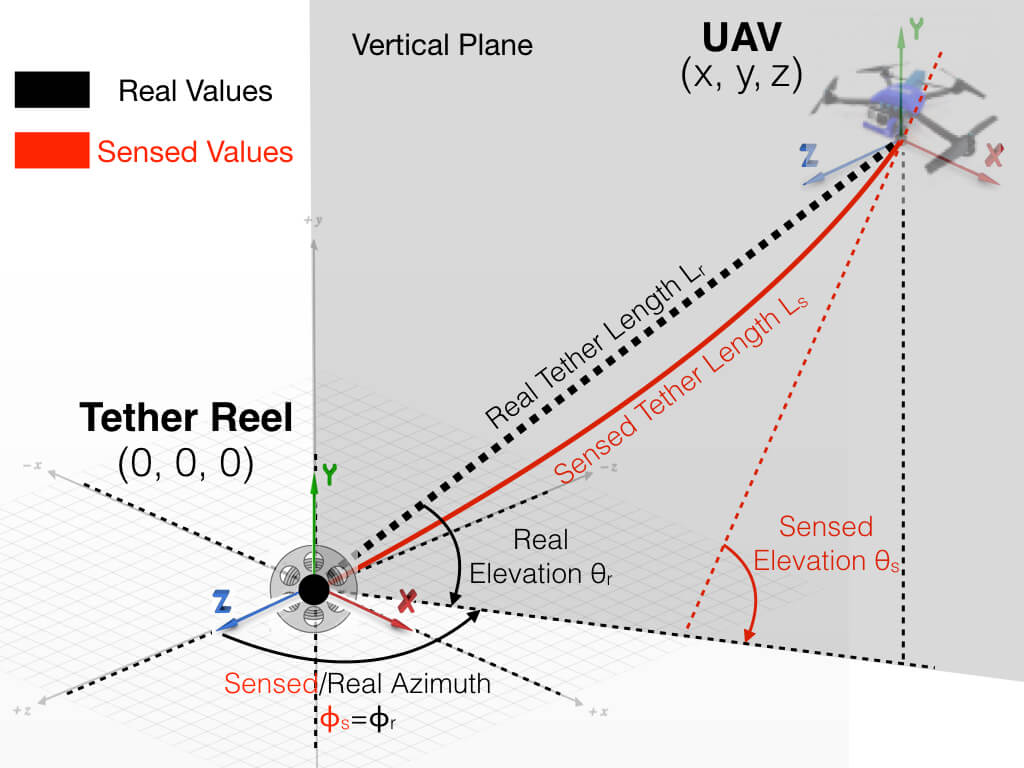}%
\label{fig::real_and_sensed}}
\subfloat[FBD of UAV]{\includegraphics[width=0.33\columnwidth]{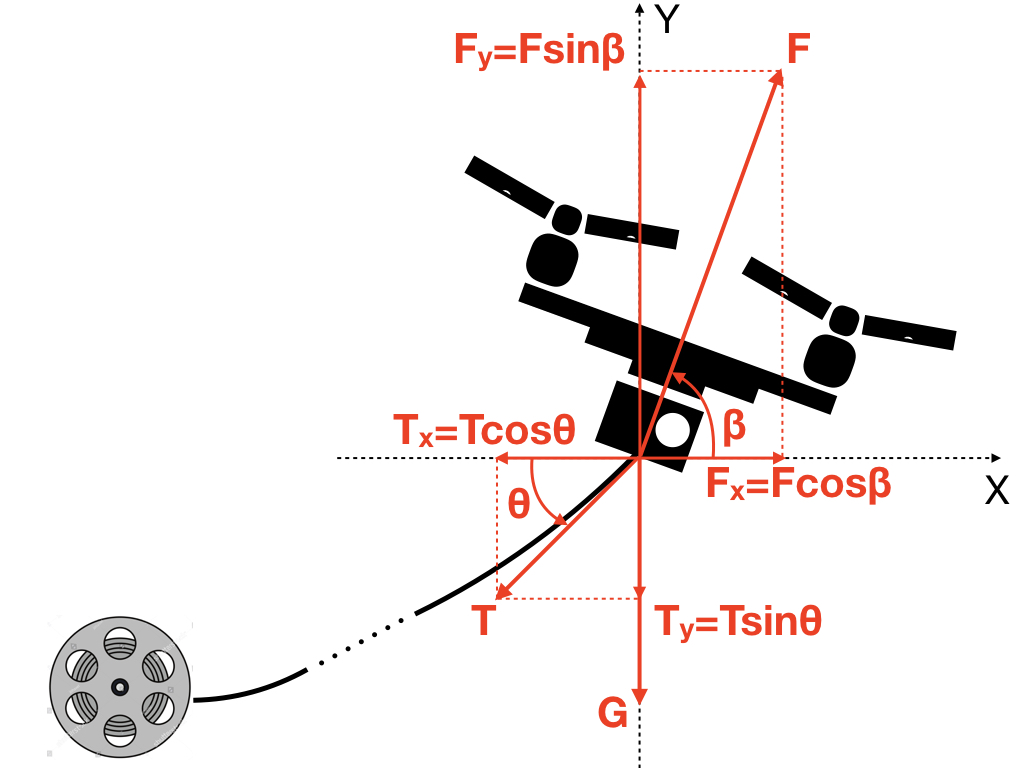}%
\label{fig::uav_fbd}}
\subfloat[FBD of Tether]{\includegraphics[width=0.33\columnwidth]{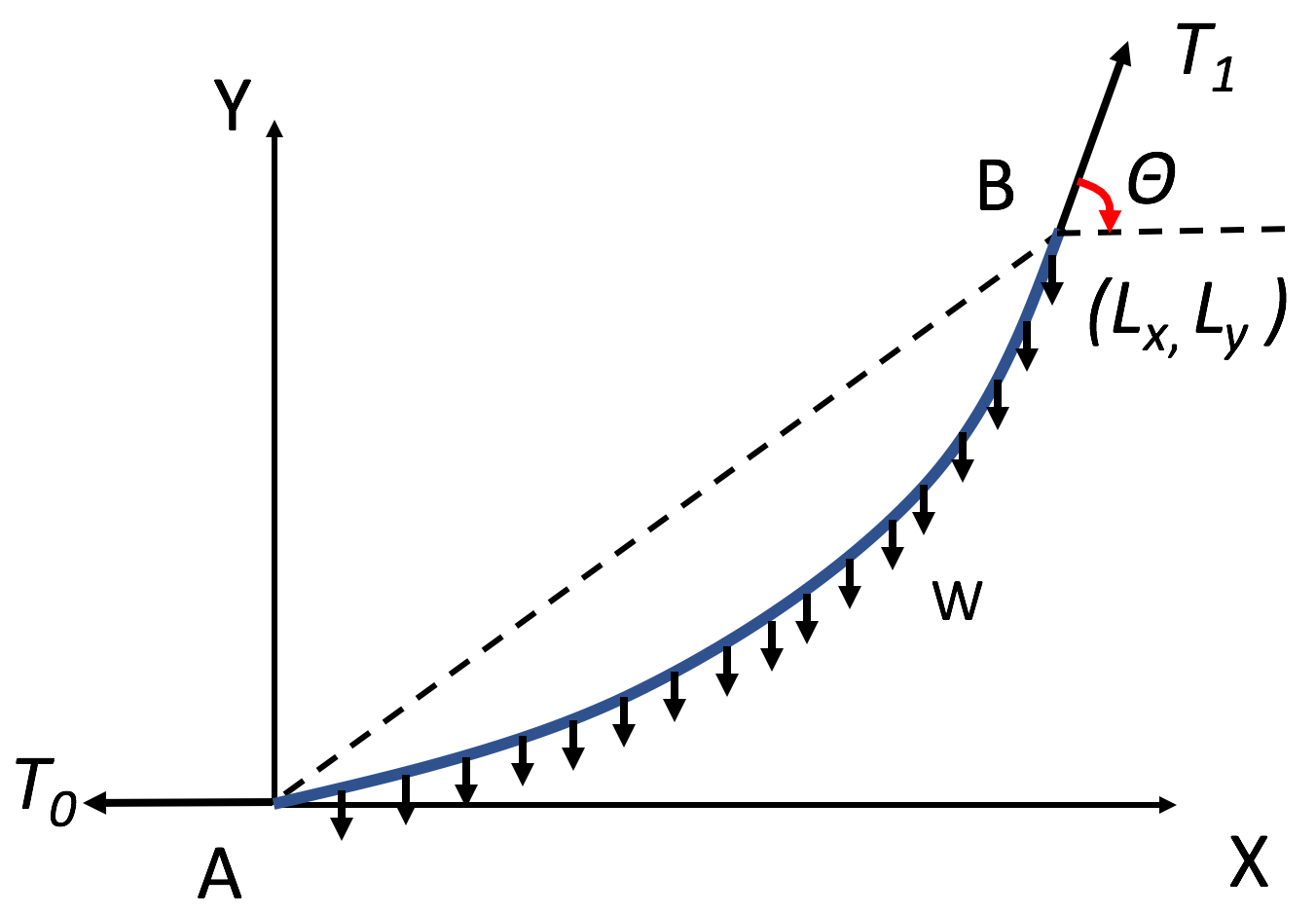}%
\label{fig::tether_fbd}}
\caption{Tether-Based Localization \cite{xiao2018indoor}}
\label{fig::localization}
\end{figure}

The aerial visual assistant takes advantage of a taut tether to localize in GPS-denied environments with a very low computational overhead. The sensory input is the tether length $L$, elevation angle $\theta$, and azimuth angle $\phi$. Although the taut tether is supposed to be straight, when the tether is long, it is pulled down by gravity, hanging as a catenary curve. Therefore a mechanics model $M$ \cite{xiao2018indoor} corrects the preliminary localization model under taut and straight tether assumption (Fig. \ref{fig::real_and_sensed}) using the Free Body Diagram (FBD) of the UAV (Fig. \ref{fig::uav_fbd}) and tether (Fig. \ref{fig::tether_fbd}) in order to achieve accurate localization of the airframe, $M: (\theta, \phi, L) \mapsto (x, y, z)$, from tether sensory input to Cartesian space location. Experiments have shown improved localization accuracy, especially with long tether \cite{xiao2018indoor}.

\subsection{Motion Primitives}
Motion primitives map the navigational goals in Cartesian space into tether-based motion commands. Two types of motion primitives are developed for the tethered aerial visual assistant: (1) Position control is based on the inverse transformation from polar to Cartesian coordinates and uses three independent Proportional–Integral–Derivative (PID) controllers to drive the position of $L$, $\theta$, and $\phi$ to their desired values (Eqn. \ref{eqn::pos_control}). (2) Velocity control based on the system's inverse Jacobian matrix computes velocity commands $L'$, $\theta'$, and $\phi'$ using an instantaneous velocity vector pointing from current location to next waypoint $d{\overrightarrow{\bf{x}}}/dt$ (Eqn. \ref{eqn::vel_control}). 

\begin{equation}
\label{eqn::pos_control}
\left\{\begin{matrix}
L =& \sqrt{x^2+y^2+z^2}\\ 
\theta =& arcsin\frac{y}{\sqrt{x^2+y^2+z^2}}\\ 
\phi = &atan2(\frac{x}{z})
\end{matrix}\right.
\end{equation}

\begin{equation}
\label{eqn::vel_control}
\begin{pmatrix}
\frac{dx}{dt}\\ 
\frac{dy}{dt}\\ 
\frac{dz}{dt}
\end{pmatrix}
=
\begin{pmatrix}
cos\theta sin\phi & -Lsin\theta sin\phi & Lcos\theta cos\phi\\ 
sin\theta & Lcos\theta & 0\\ 
cos\theta cos\phi & -Lsin\theta cos\phi & -Lcos\theta sin\phi
\end{pmatrix}
\begin{pmatrix}
L'\\ 
\theta'\\ 
\phi'
\end{pmatrix}
\end{equation}

The vehicular yaw and camera gimbal pitch are controlled using the 3-D vehicular position localization and the 3-D Cartesian coordinates of the visual assistance point of interest.  Therefore yaw and pitch of the camera viewpoint reactively point at the center of the point of interest along the entire path.  The camera gimbal roll is passively controlled by a separate Inertial Measurement Unit (IMU) to align with gravity so that the operator's viewpoint is always level to the ground. Therefore, the visual assistant's camera is pointing toward the point of interest at horizontal level along the entire motion sequence \cite{xiao2017visual}. Viewpoint drifting caused by unexpected disturbances could be fixated by an extra visual pose stabilizer \cite{dufek2017visual}. Detailed benchmarking results of the motion primitives are reported in a separate work \cite{xiao2019benchmarking}. 

\subsection{Tether Contacts Planning}
\begin{figure}
\centering
\includegraphics[width=0.6\columnwidth]{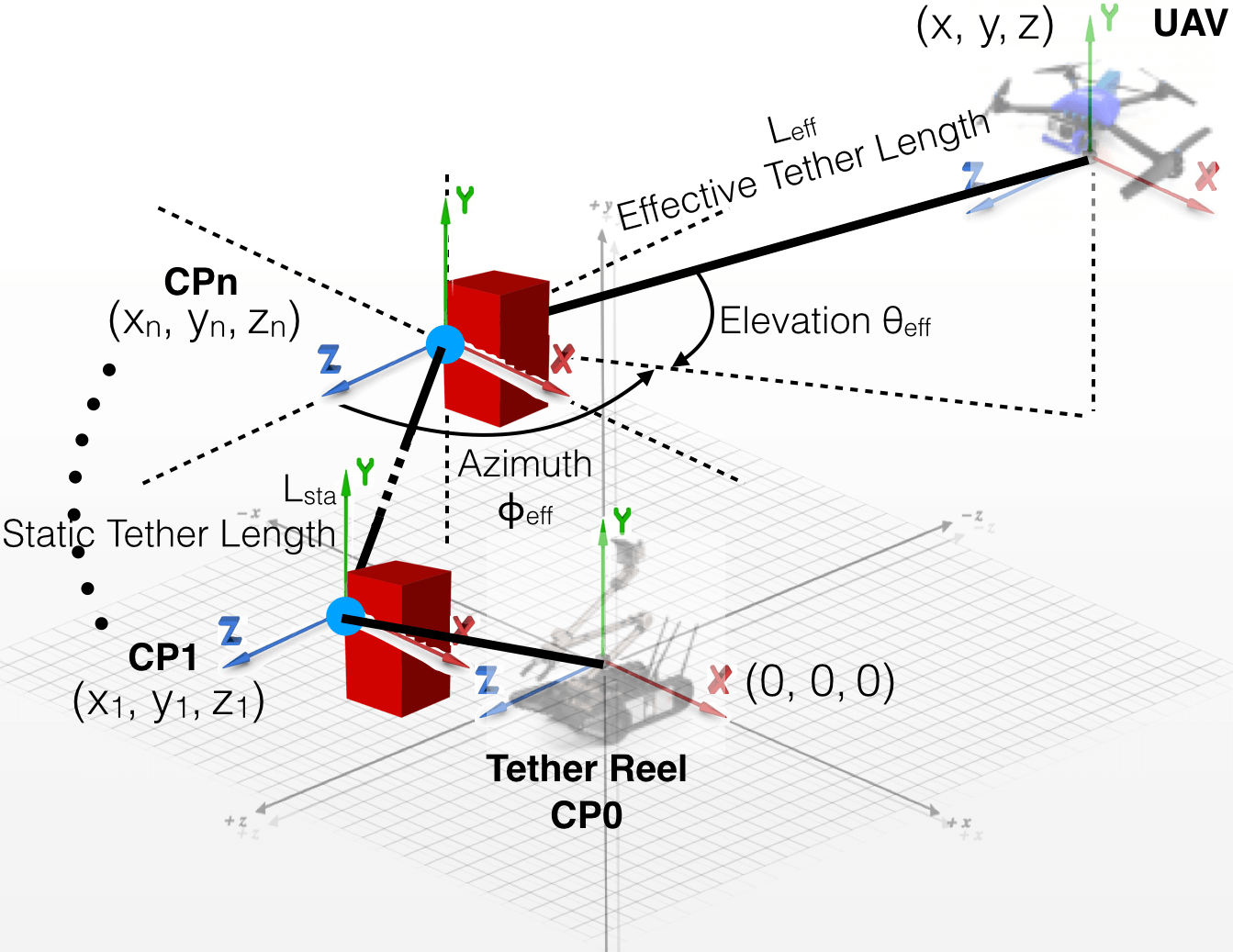}
\caption{Motion Execution with Contact(s) Planning and Relaxation}
\label{fig::contacts}
\end{figure}

In unstructured or confined environments, good viewpoints may locate behind an obstacle and the UAV cannot reach with a straight tether. In these cases, contact points of the tether with the environment are necessary to extend the reachability of the tethered visual assistant. A tether contact point(s) planning and relaxation framework \cite{xiao2018motion}, which allows the UAV to fly as if it were tetherless, is implemented on the tethered visual assistant. A new contact is planned when the current contact is no longer within line-of-sight of the UAV, while current contact is relaxed when the last contact becomes visible again. Fig. \ref{fig::contacts} shows the motion execution with multiple contact points (CPs). Given multiple contact points along the tether, static tether length denotes the portion of the tether that wraps around the obstacles, from the tether reel (CP0) to the last contact point (CPn-1) (Eqn. \ref{eqn::static_tether_length}), while the effective length is the last moving segment, from the current contact contact point (CPn) to the UAV ($x$, $y$, $z$) (Eqn. \ref{eqn::effective_controls}). Effective elevation and azimuth angles (Eqn. \ref{eqn::effective_controls}) are with respect to the last contact point (CPn), instead of the tether reel (CP0).

\begin{equation}
\label{eqn::static_tether_length}
L_{sta} =  \sum_{i=0}^{n-1}\sqrt{(x_{i+1}-x_i)^2+(y_{i+1}-y_i)^2+(z_{i+1}-z_i)^2}
\end{equation}

\begin{equation}
\label{eqn::effective_controls}
\left\{\begin{matrix}
L_{eff} = &\sqrt{(x-x_n)^2+(y-y_n)^2+(z-z_n)^2}\\
\theta_{eff} = &arcsin(\frac{y-y_n}{\sqrt{(x-x_n)^2+(y-y_n)^2+(z-z_n)^2}})\\
\phi_{eff} = &atan2(\frac{x-x_n}{z-z_n})
\end{matrix}\right.
\end{equation}

The final desired position control to the tethered aerial visual assistant is shown in Eqn. \ref{eqn::desired_controls}. For velocity control with contact point(s), the Jacobian matrix is computed based on the moving tether segment with respect to the current contact point. 

\begin{equation}
\label{eqn::desired_controls}
\left\{\begin{matrix}
L = &L_{eff} + L_{sta}\\
\theta = &\theta_{eff}\\
\phi = &\phi_{eff}
\end{matrix}\right.
\end{equation}

\section{Field Deployment}
\label{sec::system_demonstration}
The co-robots team are deployed in physical unstructured or confined environments, in an indoor mock-up and an outdoor actual after-disaster scenario. The aforementioned theories and approaches are used to realize tethered aerial visual assistance in either autonomous or tele-operated manner. This section provides detailed report of both deployments. 

\subsection{Indoor Test}
The purpose of the indoor test is to showcase the autonomous tethered aerial visual assistance. The test space is a mock-up of an indoor unstructured or confined environment, such as after-earthquake. The clutteredness is built from normal office supplies. Two main obstacles are built from cupboard boxes, which allow tether contact points to be safely formed on the edge. Therefore we enable tether contact point(s) planning in the indoor test. 

In the indoor deployment, the co-robots team drives into the affected area, with the aim of retrieving a hidden sensor among the clutter. The aerial visual assistant initially perches on the UGV's landing platform when the UGV is tele-operated to the scene. The map of the environment is given for this particular deployment or could be created by the LiDAR onboard the UGV. 

\begin{figure}
\centering
\subfloat[Entering the Scene]{\includegraphics[width=0.33\columnwidth]{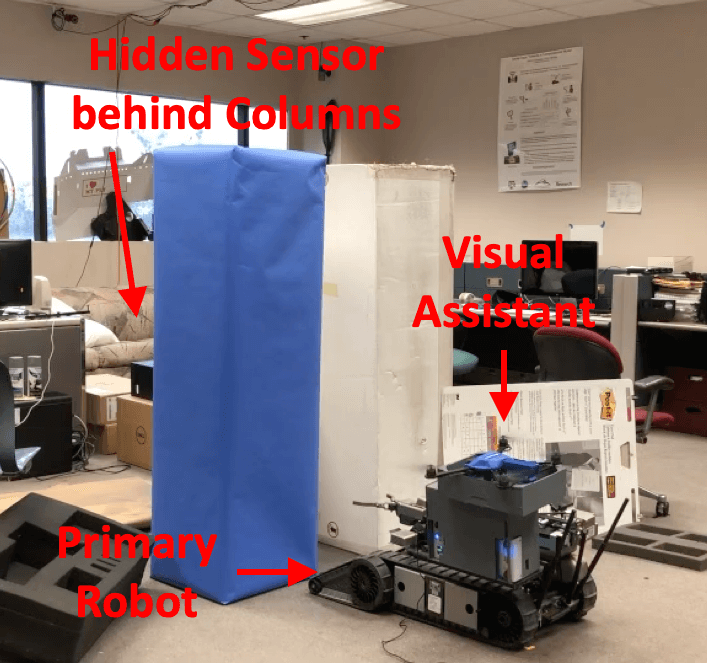}%
\label{fig::scene}}
\subfloat[Deploying for \emph{Passability}]{\includegraphics[width=0.33\columnwidth]{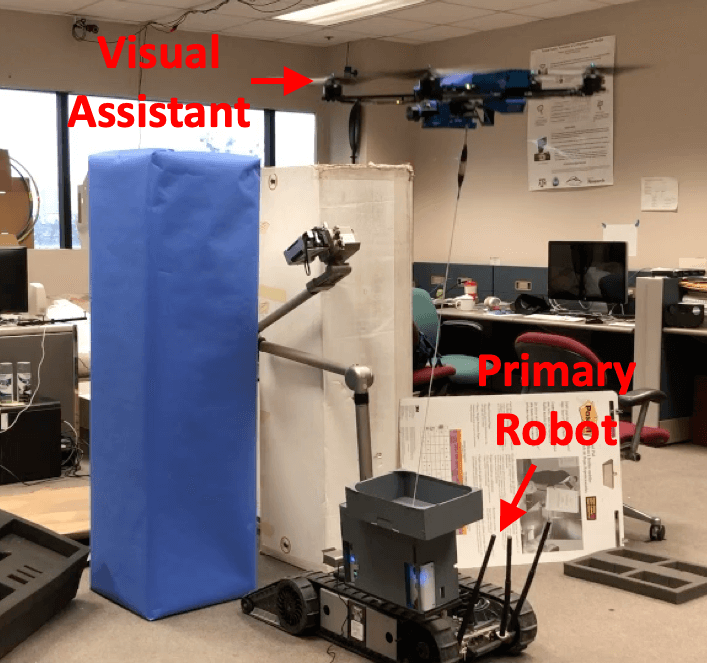}%
\label{fig::scene2}}
\subfloat[Visual Assistant View]{\includegraphics[width=0.33\columnwidth]{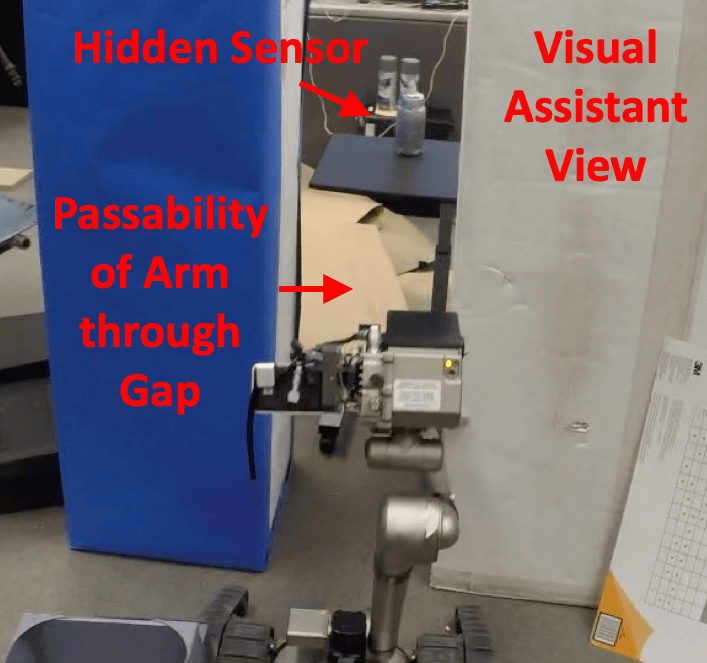}%
\label{fig::va_pass_view}}
\caption{Visual Assistance for \emph{Passability}}
\label{fig::passability_ex}
\end{figure}

The entry points to the sensor are all blocked by the clutter, leaving the only retrieval option through the narrow gap between the two column-shaped obstacles (shown in blue and white in Fig. \ref{fig::scene}). It is difficult to perform tele-operation only through the visual feedback from the onboard camera fixed to the UGV arm due to the limited field of view. Based on the viewpoint quality for \emph{passability}, the visual assistant takes off and deploys to a viewpoint from behind and above to help perceive arm \emph{passability} through gap (Fig. \ref{fig::scene2}). There is no obstacle behind the co-robots team, so a straight path is autonomously planned and executed. The visual assistant view is shown in Fig. \ref{fig::va_pass_view}, where the relative location of the arm to the narrow gap along with the hidden sensor is well perceived. 

\begin{figure}
\centering
\subfloat[Path Planning with 2 Contacts]{\includegraphics[width=0.3548\columnwidth]{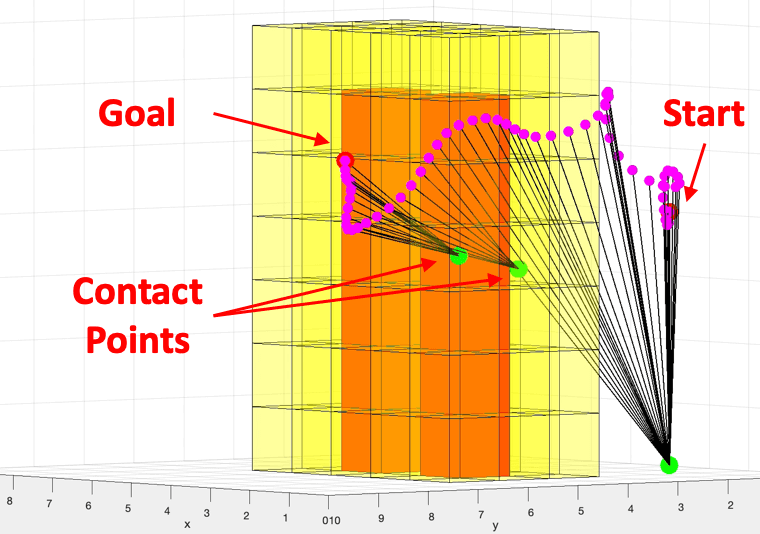}%
\label{fig::map}}
\hspace{10pt}
\subfloat[Deploying for \emph{Manipulability}]{\includegraphics[width=0.4452\columnwidth]{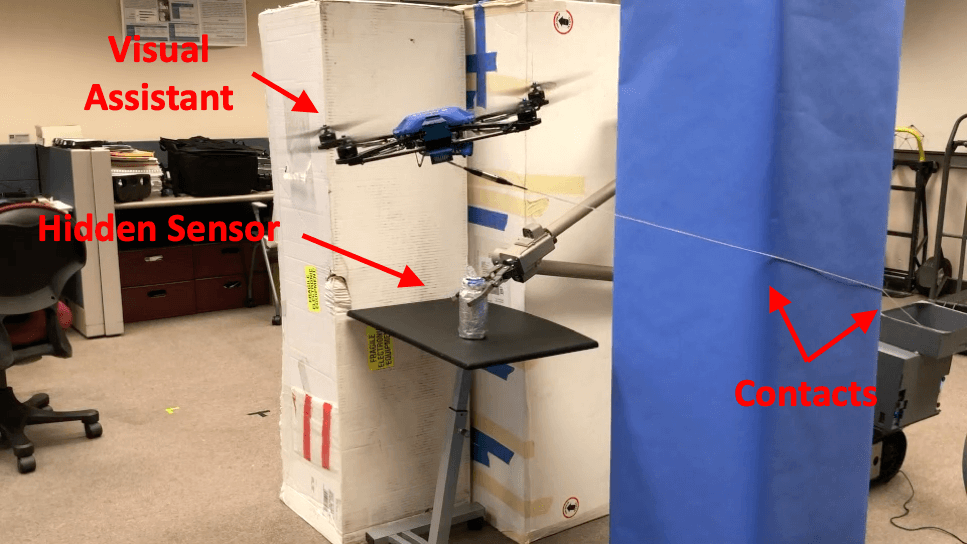}%
\label{fig::scene3}}\\
\subfloat[Onboard Camera and Visual Assistant View]{\includegraphics[width=0.5\columnwidth]{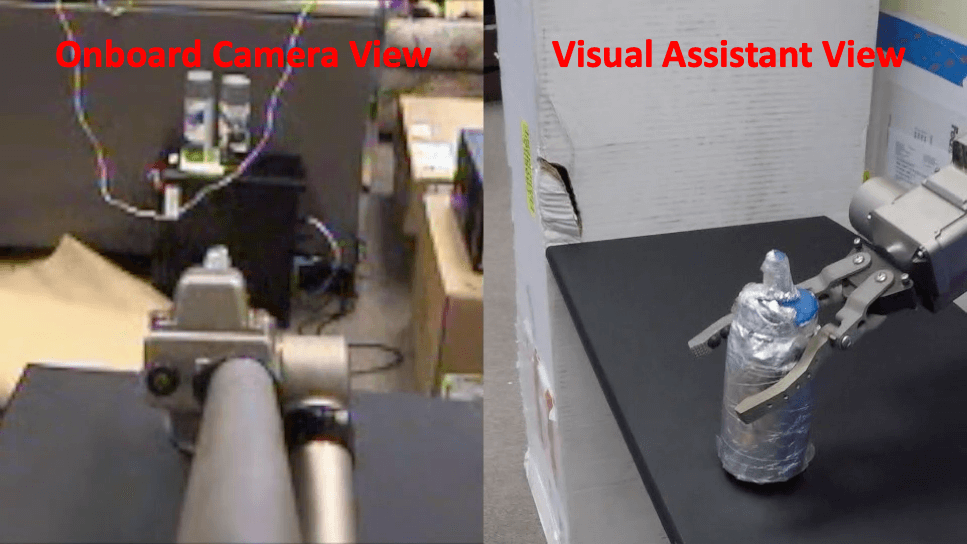}%
\label{fig::view_comparison.png}}
\caption{Visual Assistance for \emph{Manipulability}}
\label{fig::manipulability_ex}
\end{figure}

After the arm passes through the gap, the UGV needs to use the gripper at the end of the arm to grasp the sensor and retrieve it through the gap. Therefore the visual assistant switches to assisting with \emph{manipulability}. Good viewpoints for \emph{manipulability} locate at the side of the sensor and gripper. Although the path leading to the best viewpoint contains two tether contact points with the blue obstacle, the quality of the viewpoint for \emph{manipulability} makes it worth the extra risk. Fig. \ref{fig::map} shows the obstacles (red), inflated space for UAV flight tolerance (yellow), waypoints on the planned path (magenta), and two contact points on the obstacles (green, plus the tether reel as the zero-th contact point). The tether configuration is illustrated with black lines. The actual deployment is shown in Fig. \ref{fig::scene3}. Note that it is impossible to achieve sufficient situational awareness behind the obstacles using any onboard cameras on the UGV. The onboard camera view on the left of Fig. \ref{fig::view_comparison.png} completely misses the depth perception, and the sensor is even blocked entirely by the arm. With this onboard view alone, the risk of not reaching or even knocking off the sensor is high. This lack of depth perception is compensated by the visual assistant view (right). The relative distance between the sensor and the gripper is clearly perceived from this perpendicular view. It is also very easy to decide if the gripper has successfully grasped the sensor and is ready to retrieve. 

\subsection{Outdoor Test}
\begin{figure}
\centering
\subfloat[View from Entry Point]{\includegraphics[width=0.4\columnwidth]{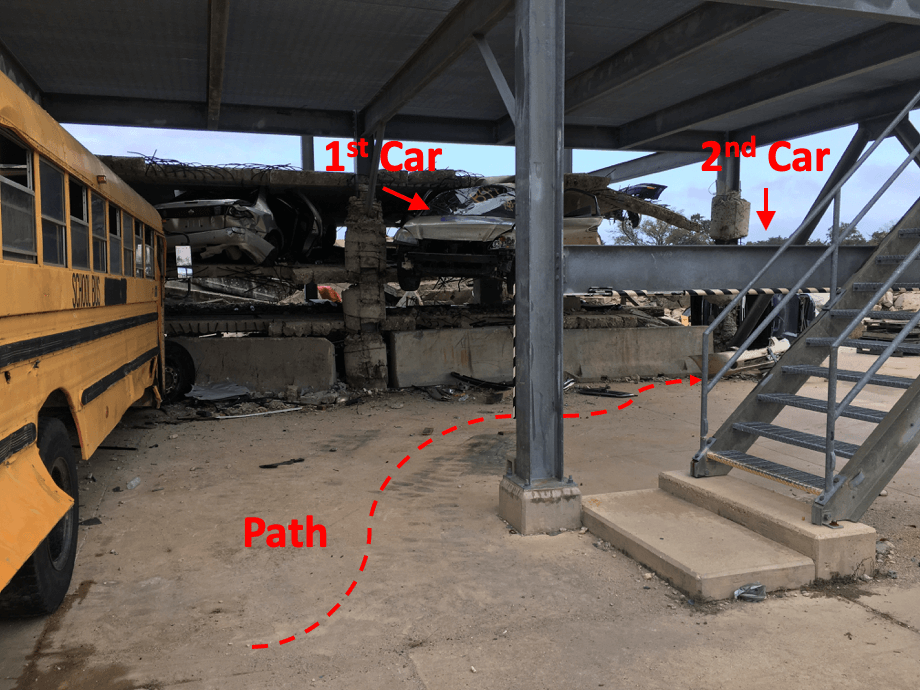}%
\label{fig::dc1}}
\hspace{10pt}
\subfloat[View from End Point]{\includegraphics[width=0.4\columnwidth]{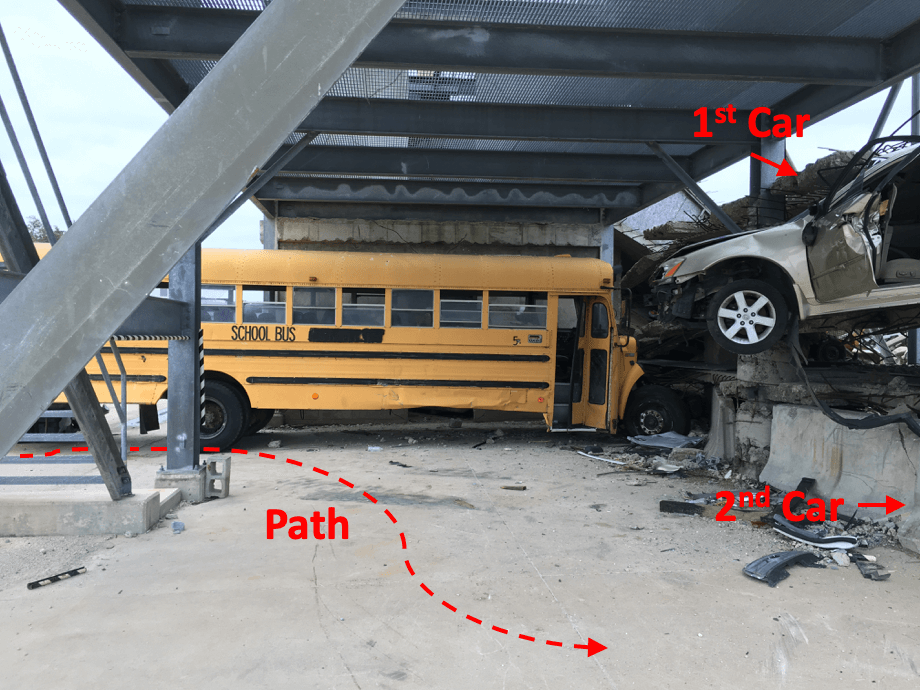}%
\label{fig::dc2}}
\caption{Disaster City \textsuperscript{\textregistered} Prop 133}
\label{fig::dc}
\end{figure}

In addition to the indoor mock-up environment, the co-robots team is also deployed in a more realistic outdoor disaster environment, Disaster City\textsuperscript{\textregistered} Prop 133 in College Station, Texas (Fig. \ref{fig::dc}). The purpose of the outdoor test is to showcase tele-operation tasks in realistic unstructured or confined environments, which were impossible for the UGV alone, but are now made possible by the visual assistance from the tethered UAV. This physical environment simulates a collapsed multi-story building and the mission for the co-robots team is to navigate into the building and search for victims and threats in two stranded cars. The first car was on the second floor before the collapse but is now squeezed down between the collapsed second and third floor. It remains unknown if there is victim trapped in the car after the collapse. The second car was tipped over during the collapse, with its sunroof open on the side. Victim may be present in the car, and possible hazardous material needs to be inspected and retrieved if necessary to prevent further incidents. To conduct this mission, the tele-operated UGV path is pre-planned by the operator, entering from the entry point, reaching the first and then second car (Fig. \ref{fig::dc}). 

\begin{figure}
\centering
\subfloat[Take-off and Deployment]{\includegraphics[width=0.4\columnwidth]{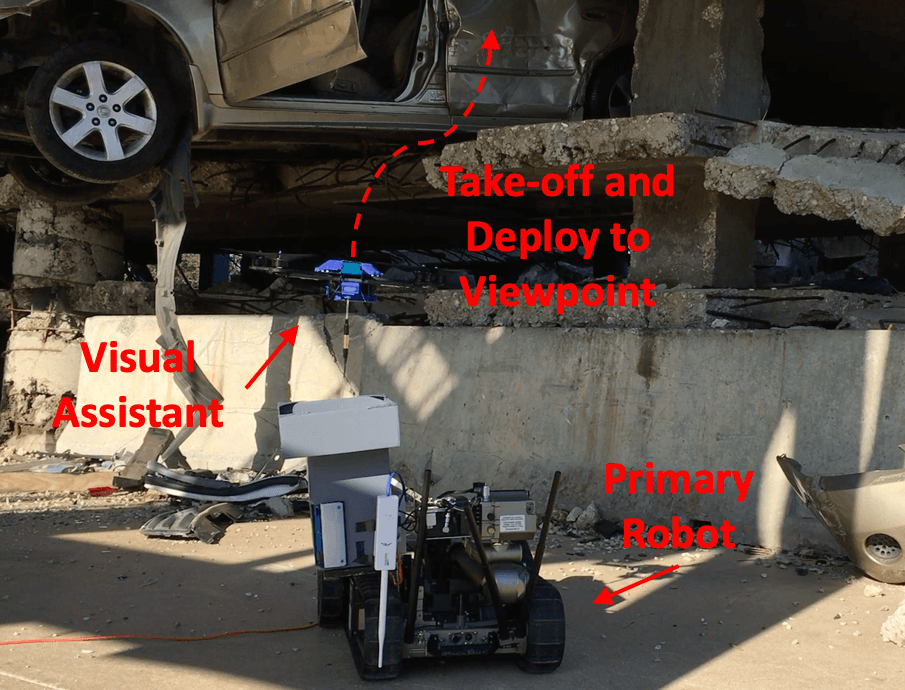}%
\label{fig::car11}}
\hspace{10pt}
\subfloat[No Victim in 1st Car]{\includegraphics[width=0.4\columnwidth]{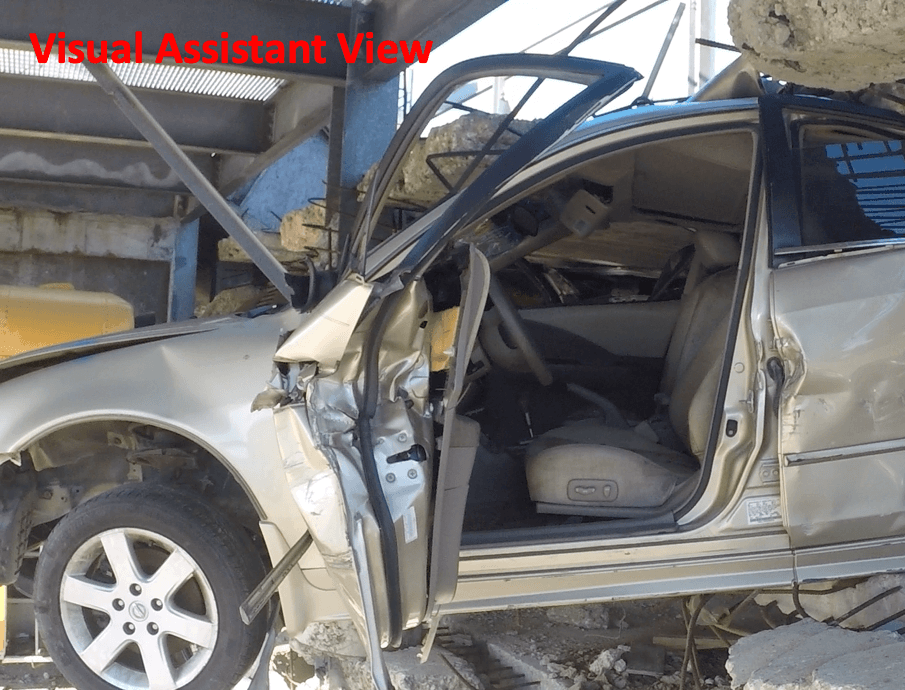}%
\label{fig::car12}}
\caption{Enhanced Coverage through Visual Assistance for 1st Car}
\label{fig::car1}
\end{figure}

The co-robots team moves as a whole during tele-operated ground locomotion, with the UAV initially perching on the landing platform on the UGV. After the team reaches the first stranded car, primary robot's onboard camera is not able to reach the height to search victims inside the car. Visual assistant therefore takes off from the landing platform and autonomously navigates to a manually specified viewpoint to look inside the car (Fig. \ref{fig::car11}). Through the elevated viewpoint provided by the visual assistant, it is confirmed that no victim is trapped in the first car. The visual assistant then lands back on the primary robot and the team is tele-operated to the second car. 

\begin{figure}
\centering
\subfloat[Inspection for 2nd Car]{\includegraphics[width=0.344\columnwidth]{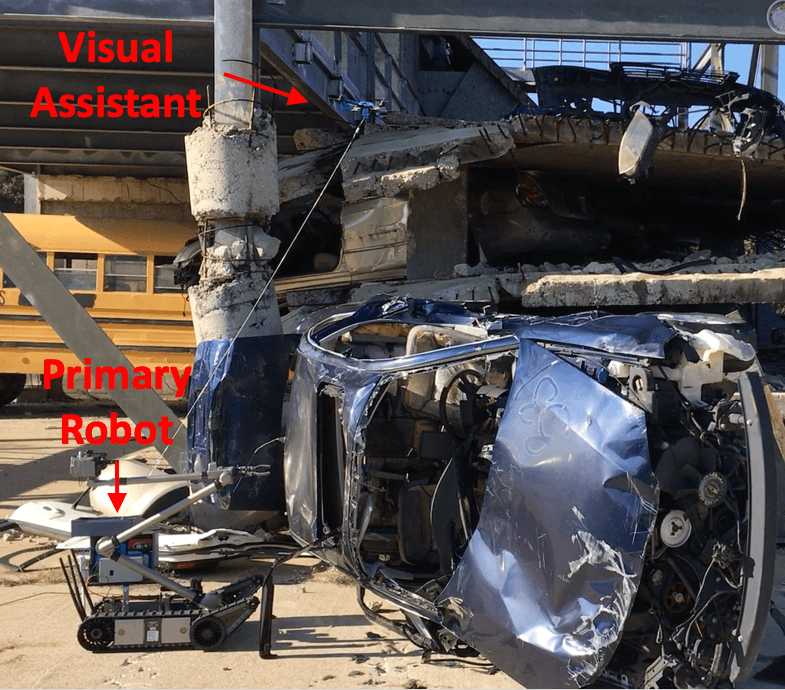}%
\label{fig::car21}}
\hspace{10pt}
\subfloat[Assisting Insertion Depth Perception]{\includegraphics[width=0.456\columnwidth]{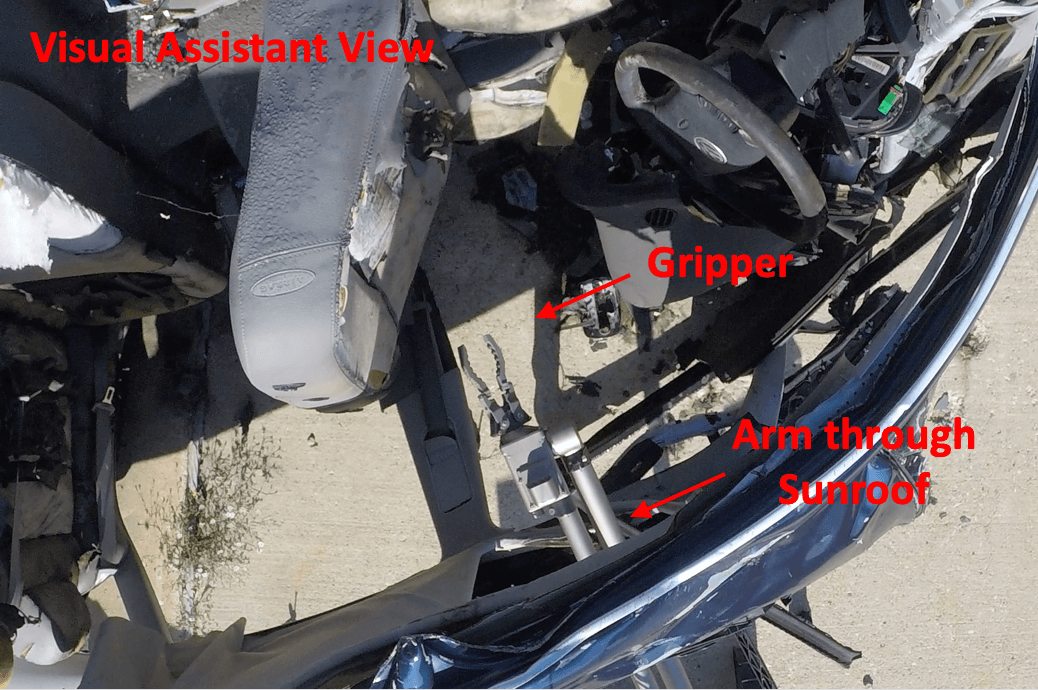}%
\label{fig::car22}}
\caption{Car Inspection through Open Sunroof for 2nd Car}
\label{fig::insertion}
\end{figure}

For the second car, the interior requires a thorough inspection. If hazardous material is present, it needs to be retrieved and disposed. The open sunroof on the side is the only possible insertion point of the UGV arm for further operations. The insertion progress is not fully perceivable through the UGV's onboard camera. Therefore the UAV is deployed. For safety reasons, tether contact points are not allowed because the sharp edges on the car may damage the tether. In this scenario, the navigational goal for the UAV is not automatically selected, but manually specified above the side window (Fig. \ref{fig::car21}). Looking down through the side window, the depth of the arm insertion into the car interior is clearly visible (Fig. \ref{fig::car22}). No victim or hazardous material exists in the car. The visual assistant lands, the co-robots team finishes the mission and navigates back. 

\section{Conclusion}
\label{sec::conclusion}
In this work, a co-robots team equipped with autonomous visual assistance using a tethered UAV for robot tele-operations in unstructured or confined environments is presented. In those DDD environments inaccessible to humans, the tele-operated primary ground robot projects human presence to and act upon remote environments, while the autonomous visual assistant provides enhanced situational awareness to the human operator. The tele-operated ground primary robot, the autonomous aerial visual assistant, and the human operator work seamlessly through a sophisticated co-robots system, interfacing from the remote field to the control center. A formal study on visual assistance viewpoint quality is performed based on the cognitive science concept of Gibsonian affordances. The formed theory is applicable not only to this particular visual assistant, but to all affordance-based tele-operation tasks. In order to navigate through unstructured or confined environments, as the use cases where visual assistance is most necessary and useful, we use a formal risk reasoning framework based on propositional logic and probability theory to enable our visual assistant's risk-awareness in those challenging spaces. This risk-reasoning framework and risk-aware planning paradigm is not limited to our particular application, but to improve robots' risk-awareness and trust-worthiness in general. This work also opens up a new regime for indoor aerial locomotion: tethered flight. A low-overhead tether-based localization technique, two motion primitives to translate Cartesian motion to tethered commands, and a tether contact point(s) planning and relaxation algorithm work together to take advantage of the tether while overcome the shortcomings brought by it. The tethered motion suite enables the tethered UAV flying as if it were tetherless. 

We are currently integrating the LiDAR mapping capability on the UGV into the entire autonomous visual assistance pipeline. We are also looking at integrating the interfaces of both the primary robot and the visual assistant into one integrated interface. Studies on tele-operation performance with visual assistance need to be carried out using physical robots, instead of in simulation. This will be done in a proof of concept high fidelity field test in which an expert operator will teleoperate PackBot in two highly realistic tasks (going through a closed door and inserting a sensor probe) in a realistic setting while being provided with a viewpoint from the presumed best manifold for each affordance and then presumed worst manifold for each affordance. 

For the tethered UAV, we have observed slight tether contact movement along the edge where the contact is formed. This could be modeled and accounted for in order to improve flight accuracy even with multiple contact points in the future. Contact point safety needs to be reasoned so that the reachability of the visual assistant could be extended even in hostile environments, such as in Disaster City\textsuperscript{\textregistered} Prop 133, by allowing the possibility of safe contact points. 

\subsubsection*{Acknowledgments}
This work is supported by NSF 1637955, NRI: A Collaborative Visual Assistant for Robot Operations in Unstructured or Confined Environments and NSF 1945105, NRI: Best Viewpoints for External Robots or Sensors Assisting Other Robots. 

\bibliographystyle{apalike}
\bibliography{jfrExampleRefs}

\end{document}